\documentclass[12pt]{article} 
\usepackage{amsmath,amsthm,amssymb,bm,mathrsfs,}
\usepackage{algorithm,algorithmic,graphicx,subfig}
\usepackage{caption,enumerate,natbib,listings,setspace}
\usepackage[OT1]{fontenc}
\usepackage[colorlinks,linkcolor=red,anchorcolor=blue,citecolor=blue]{hyperref}
\usepackage[margin=1in]{geometry}
\usepackage[protrusion=false,expansion=true]{microtype}
\usepackage[title]{appendix}
\usepackage{bbm}
\usepackage{comment}
\theoremstyle{plain}
\let\hat\widehat
\let\tilde\widetilde

\newcommand{\ab}{\mathbf{a}}

\newcommand{\eb}{\mathbf{e}}

\newcommand{\ib}{\mathbf{i}}

\newcommand{\vb}{\mathbf{v}}

\newcommand{\xb}{\mathbf{x}}

\newcommand{\be}{\bm{e}}

\newcommand{\bx}{\bm{x}}

\newcommand{\Ab}{\mathbf{A}}

\newcommand{\Ib}{\mathbf{I}}

\newcommand{\Ub}{\mathbf{U}}

\newcommand{\Xb}{\mathbf{X}}
\newcommand{\Yb}{\mathbf{Y}}

\newcommand{\bA}{\bm{A}}

\newcommand{\bX}{\bm{X}}

\newcommand{\cA}{\mathcal{A}}

\newcommand{\cD}{\mathcal{D}}
\newcommand{\cE}{\mathcal{E}}

\newcommand{\cH}{\mathcal{H}}

\newcommand{\cM}{\mathcal{M}}

\newcommand{\cO}{\mathcal{O}}

\newcommand{\cS}{{\mathcal{S}}}

\newcommand{\cX}{\mathcal{X}}
\newcommand{\cY}{\mathcal{Y}}
\newcommand{\cZ}{\mathcal{Z}}

\newcommand{\RR}{\mathbb{R}}

\newcommand{\bbeta}{\bm{\beta}}

\newcommand{\argmin}{\mathop{\mathrm{argmin}}}
\newcommand{\argmax}{\mathop{\mathrm{argmax}}}

\newcommand{\diag}{{\rm diag}}

\newtheorem{lemma}{{\bf Lemma}}

\newtheorem{theorem}{{\bf Theorem}}

\newtheorem{assumption}{{\bf Assumption}}

\newtheorem{remark}{{\bf Remark}}

\usepackage{xr}
\usepackage{color}

\newcommand{\change}[1]{{\leavevmode\color{black}{#1}}}
\definecolor{blue-violet}{rgb}{0.54, 0.17, 0.89}

\title{\Large{\textbf{Stochastic Low-rank Tensor Bandits for Multi-dimensional Online Decision Making}} } 

\author
{
Jie Zhou\thanks{Amazon, Email: jiezhoua@amazon.com},
Botao Hao\thanks{Deepmind, Email: bhao@google.com},
Zheng Wen\thanks{Deepmind, Email: zhengwen@google.com},
Jingfei Zhang\thanks{Miami Herbert Business School, University of Miami. Email: ezhang@bus.miami.edu.}
and
Will Wei Sun\thanks{Krannert School of Management, Purdue University. Email: sun244@purdue.edu. Corresponding author.}, 
}

\begin{document} 

\maketitle

\begin{abstract}
\noindent
Multi-dimensional online decision making plays a crucial role in many real applications such as online recommendation and digital marketing. In these problems, a decision at each time is a combination of choices from different types of entities. To solve it, we introduce stochastic low-rank tensor bandits, a class of bandits whose mean rewards can be represented as a low-rank tensor. We consider two settings, tensor bandits without context and tensor bandits with context. In the first setting, the platform aims to find the optimal decision with the highest expected reward, a.k.a, the largest entry of true reward tensor. In the second setting, some modes of the tensor are contexts and the rest modes are decisions, and the goal is to find the optimal decision given the contextual information. We propose two learning algorithms \texttt{tensor elimination} and \texttt{tensor epoch-greedy} for tensor bandits without context, and derive finite-time regret bounds for them. Comparing with existing competitive methods, \texttt{tensor elimination} has the best overall regret bound and \texttt{tensor epoch-greedy} has a sharper dependency on dimensions of the reward tensor. Furthermore, we develop a practically effective Bayesian algorithm called \texttt{tensor ensemble sampling} for tensor bandits with context. Extensive simulations and real analysis in online advertising data back up our theoretical findings and show that our algorithms outperform various state-of-the-art approaches that ignore the tensor low-rank structure. 
\end{abstract}


\newpage
\baselineskip=24pt

\section{Introduction}
\label{sec:introduction}
\begin{figure}[h!]
	\centering
	\includegraphics[scale=0.3]{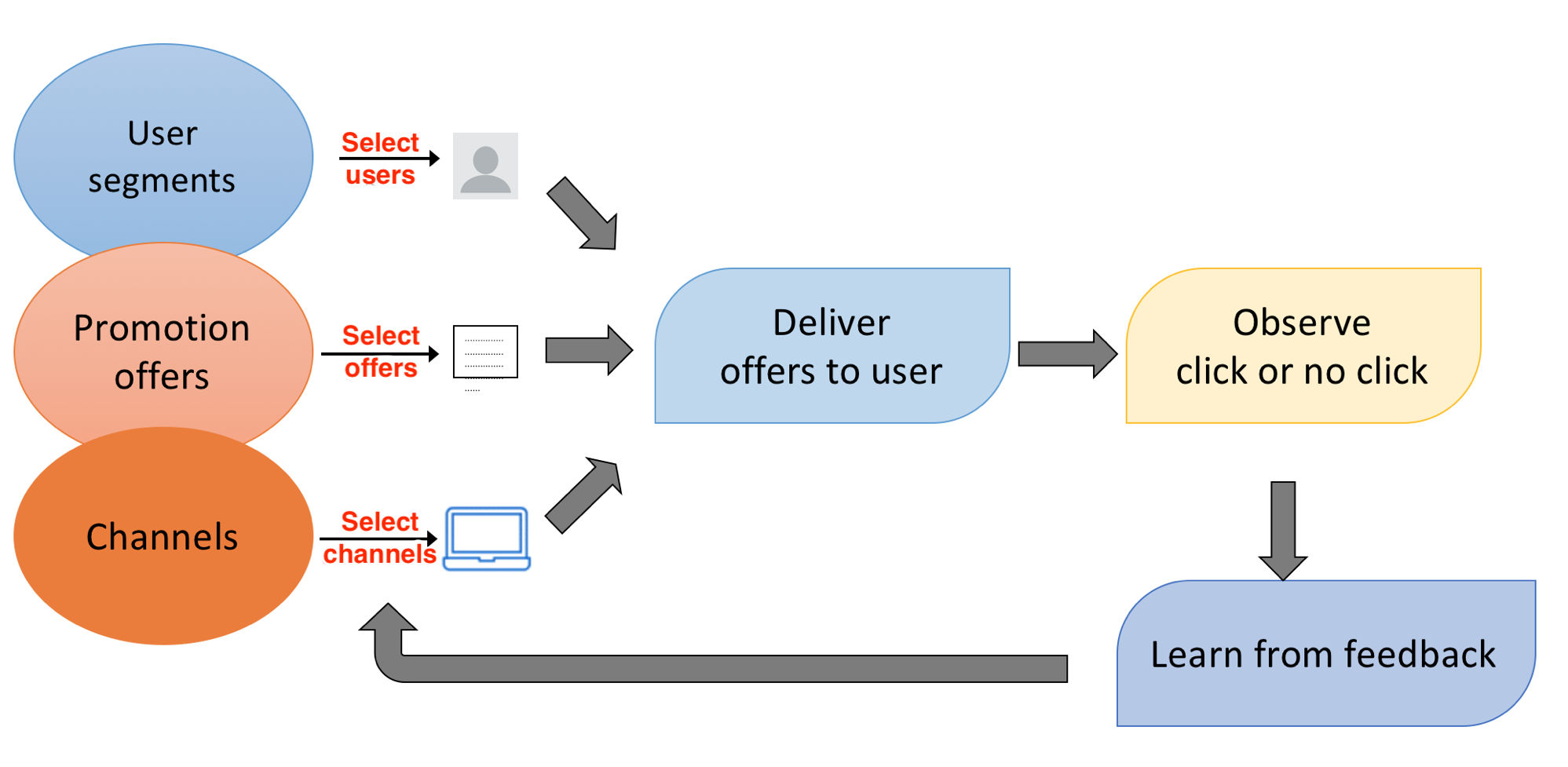}
	\includegraphics[scale=0.35]{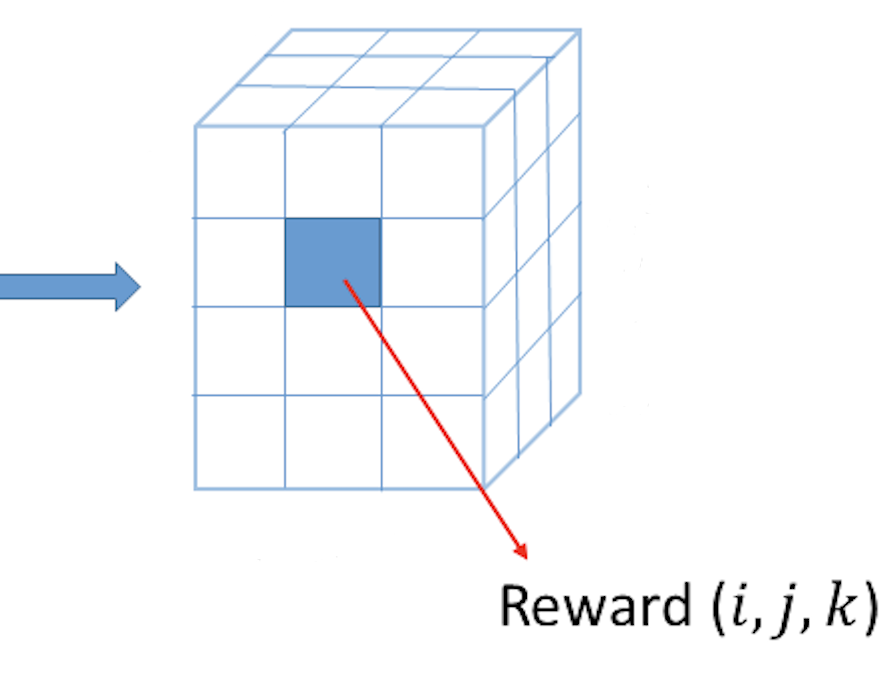}
	\caption{An example of interactive multi-dimensional online decision making. The rewards from all sequential multi-dimensional decisions formulate a tensor. }
	\label{fig:example2}
\end{figure}

The tensor, which is also called multidimensional array, is well recognized as a powerful tool to represent complex and unstructured data.
Tensor data are prevalent in a wide range of applications such as recommender systems, computer vision, bioinformatics, operations research, and etc \citep{frolov2017tensor,bi2018multilayer, song2019tensor, bi2021tensors, bi2022improving}.
The growing availability of tensor data provides a unique opportunity for decision-makers to efficiently develop multi-dimensional decisions for individuals. In this paper, we introduce tensor bandits problem where a decision, also called an arm, is a combination of choices from different entity types, and the expected rewards formulate a tensor.
The problem is motivated by numerous applications in which the agent (the platform) must recommend multiple different entity types as one arm. For example, in an advertising campaign a marketer wants to promote a new product with various promotion offers.
The goal is to choose an optimal triple user segment$\times$offer$\times$channel for this new product to boost the effectiveness of the advertising campaign.
At each time, after making an action, i.e., pulling the arm (user $i$, offer $j$, channel $k$), the leaner receives a reward, e.g., clicking status or revenue, indicating the user segment $i$'s feedback on promotion offer $j$ on marketing channel $k$. The rewards of all these three-dimensional arms formulate an order-three tensor, see Figure \ref{fig:example2} for an illustration.
Similarly, a clothing website may want to recommend the triple top$\times$bottom$\times$shoes to a user that fits the best together. Each arm is the triple of three entities. In these applications, the agent needs to pull an arm by considering multiple entities together and learn to decide which arm provides the highest reward. 

Traditional tensor methods focus on static systems where agents do not interact with the environment, and typically suffer the cold-start issue in the absence of information from new customers, new products or new contexts \citep{song2019tensor}. However, in many real applications, agents receive feedback from the environment interactively and new subjects enter the system sequentially. See Figure \ref{fig:example2} for an illustration of such interactive sequential decision making. In each round, the agent recommends a promotion offer to a chosen user segment in a channel, and then the agent receives a feedback from this user segment. Based on this instant feedback, the agent needs to update the model to improve the user targeting accuracy in the future.

Bandit problems are basic instances of interactive sequential decision making and now play an important role in vast applications such as revenue management, online advertising, and recommender system \citep{li2010contextual, bubeck2012, lattimore2020bandit}. In bandit problems, at each time step the agent chooses an arm/action from a list of choices based on the action-reward pairs observed so far, and receives a random reward that is conditionally independently drawn from the unknown reward distribution given the chosen action. The objective is to learn the optimal arm that maximizes the sum of the expected rewards. The heart of bandit problems is to address the fundamental trade-off between exploration and exploitation in sequential experiments. At each time step, after receiving the feedback from users, the agent faces a decision dilemma. The agent can either exploit the current estimates to optimize decisions or explore new arms to improve the estimates and achieve higher payoffs in the future. Our considered tensor bandits problem can be viewed as a higher-order extension of the standard bandit problem, which generalizes a scalar arm to a multi-dimensional arm and correspondingly generalizes a vector reward to a tensor reward case. 

In this article, we introduce stochastic low-rank tensor bandits for multi-dimensional online decision-making problems. These are a class of bandits whose mean rewards can be represented as a low-rank tensor and arms are selected from different entity types. 
\change{The assumption of low-rankness is well-adopted in the literature on tensors. It effectively reduces model complexity and finds widespread applications in practical scenarios such as online recommendation systems and digital marketing \citep{Sun2016, bi2021tensors, ide2022targeted}. More practical justifications for the use of low-rank tensors can be found in the survey paper \citet{song2019tensor}.} To balance the exploration-exploitation trade-off, we propose two algorithms for tensor bandits, \texttt{tensor epoch-greedy} and \texttt{tensor elimination}. The \texttt{tensor epoch-greedy} proceeds in epochs, with each epoch consisting of an exploration phase and an exploitation phase. In the exploration phase, arms are randomly selected and in the exploitation phase, arms that expect the highest reward are pulled. The number of steps in each exploitation phase increases with number of epochs, guided by the fact that, as the number of epochs increases, the estimation accuracy of the true reward improves and more exploitation steps are desirable. For \texttt{tensor elimination}, we incorporate the low-rank structure of reward tensor to transform the tensor bandit into linear bandit problem with low-dimension and then employ the upper confidence band (UCB) \citep{lai1985asymptotically} to enable the uncertainty quantification. The UCB has been very successful in bandit problems, leading to an extensive literature on UCB algorithms for standard multi-armed bandits \citep{lattimore2020bandit}. However, employing the successful UCB strategy in low-rank tensor bandits encounters a critical challenge, as the tensor decomposition is a non-convex problem. When the data is not uniformly randomly collected but adaptively collected, the concentration results for the low-rank tensor components remain elusive thus far. Our \texttt{tensor elimination} approach considers a tensor spectral-based rotation strategy that preserves the tensor low-rank information and meanwhile enables uncertainty quantification. 


In addition to these methodological contributions, in theory we further derive the finite-time regret bounds of our proposed algorithms and show the improvement over existing methods. Low-rank tensor structure has imposed fundamental challenges, as the proof strategies for existing bandit algorithms are not directly applicable to our tensor bandits problem. So the regret analysis of tensor bandits demands new technical tools. In theory, we show that two existing competitors: (1) \texttt{vectorized UCB} which vectorizes the reward tensor into a vector and then applies UCB \citep{auer2002}; and (2) \texttt{matricized ESTR} which unfolds the reward tensor into a matrix and then applies matrix bandit ESTR \citep{jun2019bilinear}, both lead to sub-optimal regret bounds. Table \ref{table:com} illustrates the comparison of our regret bounds and the regret bounds of these two competitors. \change{Importantly, we prove that \texttt{tensor epoch-greedy} has better dependency on tensor dimensions and worse dependency on time horizon compared with the other methods. Therefore, it has superiority over other methods in two scenarios: (1) when the time horizon is short, e.g., the market campaign has a small time budget; or (2) when the dimensions are high. In contrast, the regret bound of \texttt{tensor elimination} is always better than the two existing competitors due to its sharper dependency on the dimensions, and also has advantages over \texttt{tensor epoch-greedy} when time horizon is long since it has better dependency on time horizon. These theoretical guarantees and insights are important as they help us better understand the algorithms and when one might be preferred over the other.}

\begin{table}
\centering
\begin{tabular}{ |c|c| } 
\hline
Algorithm & Regret bound \\ 
\hline
\texttt{tensor epoch-greedy} & $\tilde{\cO}(p^{d/2} + p^{(d+1)/3}n^{2/3})$\\
\hline
\texttt{tensor elimination} &  $\tilde{\cO}(p^{d/2}+ p^{(d-1)/2}n^{1/2})$\\
\hline
\texttt{vectorized UCB} &  $\tilde{\cO}(p^d+ p^{d/2}n^{1/2})$\\
\hline
\texttt{matricized ESTR} &  $\tilde{\cO}(p^{d-1}+ p^{3(d-1)/2}n^{1/2})$\\
\hline
\end{tabular}
\caption{\label{table:com}Regret bounds of our proposed \texttt{tensor epoch-greedy} and \texttt{tensor elimination}, as well as the competitors \texttt{vectorized UCB} and \texttt{matricized ESTR}. Here $n$ denotes the time horizon, $p = \max\{p_1,\ldots,p_d\}$ denotes the maximum tensor dimension and $d$ denotes the order of the reward tensor. We consider $d\ge 3$, the maximum tensor rank $r=\cO(1)$, and use $\tilde{\cO}$ to denote $\cO$ ignoring logarithmic factors.}
\end{table}


Finally, we consider an interesting extension of tensor bandits when the contextual information is available. In the aforementioned tensor bandits setting, the goal is to find the optimal arm corresponding to the largest entry of the reward tensor. This setting is called tensor bandits without context. When some modes of the reward tensor are contextual information, we encounter contextual tensor bandits. Take the online advertising data considered in Section \ref{sec:real} as an example. Users use the online platform on some day of the week, and the platform can only decide which advertisement to show to this given user at the given time. In this example, the user mode and the day-of-week mode of the reward tensor are both contextual information and both are not decided by the platform. This is the key difference to the user targeting example shown in Figure \ref{fig:example2}. Because of this, many of the aforementioned methods are no longer applicable. In this paper we further develop \texttt{tensor ensemble sampling} for contextual tensor bandits that utilizes Thompson sampling \citep{russo2018tutorial} and ensemble sampling \citep{lu2017}. Thompson sampling is a powerful Bayesian algorithm that can be used to address a wide range of online decision problems. The algorithm, in its basic form, first initializes a prior distribution over model parameters, and then samples from its posterior distribution calculated using past observations. Finally, an action is made to maximize the reward given the sampled parameters. The posterior distribution can be derived in closed-form in a few special cases such as the Bernoulli bandit \citep{russo2018tutorial}. With more complex models such as our low-rank tensor bandit problem, the exact calculation of the posterior distribution becomes intractable. In this case, we consider an ensemble sampling approach \citep{lu2017} that aims to approximate Thompson sampling while maintaining computational tractability. In an online advertising application, our \texttt{tensor ensemble sampling} is empirically successful and reduces the cumulative regret by 75\% compared to the benchmark methods.

There are several lines of research that are related to but also clearly distinctive of the problem we address. The first line is tensor completion \citep{yuan2016tensor, song2019tensor, zhang2019cross, cai2021nonconvex, xia2021statistically, han2022optimal}. While we employ similar low-dimensional structures as tensor completion, the two problems have fundamental difference. First, a key assumption in existing tensor completion is to assume the observed entries are collected uniformly and randomly (the only exception is \cite{zhang2019cross} which assumes a special cross structure of the missing mechanism). This is largely different from our interactive online decision problem where the observed entries are collected adaptively based on some bandit policy. The difference is analogous to that between linear regression and linear bandit \citep{lattimore2020bandit}. Second, the goal of existing tensor completion is to predict all missing entries while the goal of tensor bandits is to find the {\it largest} entry in the reward tensor so that the cumulative regret is minimized. Third, these tensor completion algorithms are developed for off-line settings where data are collected all at once. They are not applicable to our online decision problem where data enter the system sequentially. On the other hand, existing online tensor completion \citep{yu2015accelerated, ahn2021accurate} for streaming data could not handle our interactive decision problem due to their uniform and random missing mechanism and non-interaction nature.

The second line of related work is low-rank matrix bandit. There are some works considering special rank-1 matrix bandits \citep{katariya2017stochastic, katariya2017bernoulli, trinh2019solving}. To find the largest entry of a non-negative rank-1 matrix, one just needs to identify the largest values of the left-singular and right-singular vectors. However, this is no longer applicable for higher-rank matrices. For general low-rank matrix bandits, \cite{kveton2017stochastic} handled low-rank matrix bandits but imposed strong ``hott topics" assumptions on the mean reward matrix. They assumed all rows of decomposed factor matrix can be written as a convex combination of a subset of rows. \cite{sen2017contextual} considered low-rank matrix bandits with one dimension choosing by the nature and the other dimension choosing by the agent. They derived a logarithmic regret under a constant gap assumption. However the gap may not be specified in advance. \cite{lu2018efficient} utilized ensemble sampling for low-rank matrix bandits but did not provide any regret guarantee due to the theoretical challenges in handling sampling-based exploration. \cite{jun2019bilinear} proposed a bilinear bandit that can be viewed as a contextual low-rank matrix bandit. However, their regret bound becomes sub-optimal in the context-free setting due to the use of LinUCB \citep{abbasi2011improved} for linear bandits with finitely many arms. In addition, our theory shows that unfolding reward tensor into matrix and then applying algorithm proposed by \cite{jun2019bilinear} leads to a suboptimal regret bound. \cite{lu2021low} further generalized \cite{jun2019bilinear} to a low-rank generalized linear bandit. To the best of our knowledge, there is no existing work that systematically studies tensor bandits problem. Low-rank tensor structure has imposed fundamental challenges. It is well known that many efficient tools for matrix data, such as nuclear norm minimization or singular value decomposition, cannot be simply extended to tensor framework \citep{richard2014statistical,yuan2016tensor,friedland2017nuclear, zhang2018tensor}. Hence existing algorithms and proof strategies for linear bandits or matrix bandits are not directly applicable to our tensor bandits problem. Our proposed algorithms and their regret analysis demand new technical tools.


The rest of the paper is organized as follows. Section \ref{sec:notation} reviews some notation and tensor algebra. Section \ref{sec:sec3} presents our model, two main algorithms and their theoretical analysis for the tensor bandits. Section \ref{sec:contextual_bandit} considers the extension to the contextual tensor bandits. Section \ref{sec:simulation} contains a series of simulation studies. Section \ref{sec:real} applies our algorithms to an online advertising application to illustrate their practical advantages. \change{All proofs, an analysis of approximate Thompson sampling, and additional implementation details in the experiments are included in the supplemental material. }


\section{Notation and Tensor Algebra}
\label{sec:notation}
A \textit{tensor} is a multidimensional array and the \textit{order} of a tensor is the number of dimensions it has, also referred to as the \textit{mode}.
We denote vectors using lower-case bold letters (e.g., $\xb$), matrices using upper-case bold letters (e.g., $\Xb$), and high-order tensors using upper-case bold script letters (e.g., $\cX$).
We denote the cardinality of a set by $|\cdot|$ and write $[k]=\{1,2,\ldots,k\}$ for an integer $k\geq 1$. 
For a positive scalar $x$, let $\lceil  x\rceil = \min\{z\in\mathbb N^{+}:z\geq x\}$. 
We use $\eb_j\in\mathbb R^p$ to denote a basis vector that takes 1 as its $j$-th entry and 0 otherwise. 
For a vector $\ab\in\mathbb R^d$ and $s_1\le s_2\in[d]$, let $\ab_{s_1:s_2}$ be the sub-vector $(\ab_{s_1},  \ab_{s_1+1}, \ldots, \ab_{s_2})$. 
For an order-$d$ tensor $\cX\in \mathbb R^{p_1\times \cdots\times p_d}$, define its mode-$j$ fibers as the $p_j$-dimensional vectors $\cX_{i_1,\ldots, i_{j-1}, \cdot, i_{j+1}, \ldots, i_d}$, and its mode-$j$ matricization as $\mathcal{M}_j(\cX)\in\mathbb R^{p_j\times (p_1\cdots p_{j-1}p_{j+1}\cdots p_d)}$, where the column vectors of $\mathcal{M}_j(\cX)$ are the mode-$j$ fibers of $\cX$.
For instance, for an order-3 tensor $\cX\in\mathbb R^{p_1\times p_2\times p_3}$, its mode-1 matricization $\mathcal{M}_1(\cX)\in\mathbb R^{p_1\times (p_2p_3)}$ is defined as, for $i\in[p_1], j\in[p_2], k\in[p_3]$,
\begin{equation}\label{eqn:matri}
\left[\mathcal{M}_1(\cX)\right]_{i, (j-1)p_3 + k}=\cX_{i,j,k}.
\end{equation}
For a tensor $\cX\in \mathbb R^{p_1\times p_2\times \cdots\times p_d}$ and a matrix $\Yb\in\mathbb R^{r_1\times p_1}$, we define the marginal multiplication $\cX \times_1 \Yb\in\RR^{r_1\times p_2\times \cdots \times p_d}$ as
\begin{equation}\label{eqn:marginal}
\cX\times_1 \Yb=\Big(\sum_{i_1'=1}^{p_1} \cX_{i_1',i_2,\ldots, i_d}\Yb_{i_1 ,i_1'}\Big)_{i_1\in[r_1],i_2\in[p_2] \ldots, i_d\in[p_d]}. 
\end{equation}
Marginal multiplications along other modes, i.e., $\times_2, \ldots, \times_d$, can be defined similarly. 
For $\cX, \cY\in \mathbb R^{p_1\times \cdots\times p_d}$, define the tensor inner product as 
$
\langle \cX, \cY\rangle =\sum_{i_1\in[p_1], \ldots, i_d\in[p_d]} \cX_{i_1,\ldots ,i_d} \cY_{i_1,\ldots ,i_d}.
$
The tensor Frobenius norm is defined  as
$
\|\cX\|_{\rm F}=\sqrt{\langle \cX, \cX\rangle}$, and the element-wise tensor max norm is defined as $ \|\cX\|_{\infty} = \max_{i_1, \ldots, i_d} |\cX_{i_1,\ldots, i_d}|$. 

Consider again an order-$d$ tensor $\cX\in \mathbb R^{p_1\times \cdots\times p_d}$. Letting $r_j$ be the rank of matrix $\mathcal{M}_j(\cX)$, $j\in[d]$, the tensor Tucker rank of $\cX$ is the $d$-tuple $(r_1, \ldots, r_d)$. Let $\Ub_1\in \mathbb R^{p_1\times r_1}, \ldots, \Ub_d\in \mathbb R^{p_d\times r_d}$ be the matrices whose columns are the left singular vectors of $\mathcal{M}_1(\cX), \ldots, \mathcal{M}_d(\cX)$, respectively. Then, there exists a core tensor $\cS\in \mathbb{R}^{r_1\times \cdots\times r_d}$ such that
$$
\cX = \cS \times_1 \Ub_1 \times_2\cdots \times_d \Ub_d,
$$
or equivalently, $\cX_{i_1,\ldots, i_d} = \sum_{i'_1\in[r_1], \ldots, i_d'\in[r_d]} S_{i_1',\ldots, i_d'}[\Ub_{1}]_{i_1, i_1'}\cdots[\Ub_{d}]_{i_d, i_d'}$. The above decomposition is often referred to as the tensor Tucker decomposition \citep{kolda2009}. 


\section{Tensor Bandits}
\label{sec:sec3}
In this section, we first introduce tensor bandits, followed by two new algorithms -- \texttt{tensor elimination} and \texttt{tensor epoch-greedy}. 
We then establish the finite-time regret bounds of these two algorithms, which reveal their different performances under different rate conditions and provide a useful guidance for their implementations in practice.

In tensor bandits problem, the agent interacts with an environment for $n$ time steps, and at each step, the agent faces a $d$-dimensional decision, indexed by $[p_1]\times\cdots\times [p_d]$. 
A standard multi-armed bandit can be regarded as a special case of tensor bandits with $d=1$.
At step $t\in [n]$ and given past interactions, the agent pulls an arm $I_t$, which denotes a $d$-tuple $(i_{1,t}, \ldots,i_{d,t})\in [p_1]\times \ldots\times [p_d]$. 
Correspondingly, the agent observes a reward $y_{t}\in\mathbb{R}$, drawn from a probability distribution associated with the arm $I_t$. 
Specifically, denoting the true reward tensor as $\cX\in\mathbb R^{p_1\times \ldots\times p_d}$, the agent at time $t$ receives a noisy reward 
\begin{equation}\label{eqn:tensor_bandit}
y_t = \langle \cX, \cA_t\rangle + \epsilon_t, \ \text{with} \ \cA_t = \eb_{i_{1,t}}\circ\cdots \circ \eb_{i_{d,t}},
\end{equation}
where ``$\circ$" denotes the vector outer product, $\eb_{i_{j,t}}\in\mathbb R^{p_j}$ is a basis vector, $j\in[d]$, and $\cA_t$ is a tensor indicating the location of the arm $I_t$.
For example, if the agent pulls $I_t=(i_{1,t}, \ldots,i_{d,t})$, then the $(i_{1,t}, \ldots,i_{d,t})$-th entry of $\cA_t$ is 1 while all other entries are 0. 
In \eqref{eqn:tensor_bandit}, $\epsilon_t$ is a random noise term, assumed to be sub-Gaussian in Assumption \ref{con:noise}.

The goal of our work, aligned with the central task in bandit problems, is to strike the right balance between exploration and exploitation, and to minimize the cumulative regret. 
Let the arm with the maximum true reward be 
$$
(i_{1}^*, \ldots, i_{d}^*)= \argmax_{i_{1}\in[p_1], \ldots, i_{d}\in[p_d] }\langle \cX, \eb_{i_{1}}\circ\cdots \circ \eb_{i_{d}} \rangle
$$ 
and correspondingly, denote $\cA^* = \eb_{i_{1}^*}\circ\cdots \circ \eb_{i_{d}^*}$. 
Our objective is to minimize the cumulative regret \citep{audibert2009exploration}, defined as 
\begin{equation}\label{eqn:pseduo_regret}
\textrm{R}_n=\sum_{t=1}^n\langle \cX, \cA^* \rangle-\sum_{t=1}^n \langle \cX, \cA_t\rangle.
\end{equation}
Naturally, at each step $t\in[n]$, the agent faces an exploitation-exploration dilemma, in that the agent can either choose the arm that expects the highest reward based on historical data (exploitation), so as to reduce immediate regret, or choose some under-explored arms to gather information about their associated reward (exploration), so as to reduce future regret.

At first glance, the tensor bandit problem posed in \eqref{eqn:tensor_bandit}-\eqref{eqn:pseduo_regret} can be re-formulated, via vectorization, as a standard multi-armed bandit problem of dimension $p_1\times\ldots\times p_d$. 
However, applying the existing algorithms for standard multi-armed bandits to vectorized tensor bandits may be inappropriate due to several reasons. 
First, the majority of existing solutions for multi-armed bandits require a proper initialization phase where each arm is pulled at least once, in order to give a well-defined solution \citep{auer2002finite}. 
For tensor bandits, such an initialization step can be computationally expensive or even infeasible, especially when $p_1\times\ldots\times p_d$ is large.
Second, the vectorization approach may result in a severe loss of information, as the intrinsic structures (e.g., low-rank) of tensors are largely ignored after vectorization.
Indeed, as commonly considered in recommendation systems and other applications \citep{kolda2009, Allen12,jain2014provable, bi2018multilayer, song2019tensor, xia2021statistically, bi2021tensors}, tensor objects usually have a low-rank structure and can be represented in a lower-dimensional space.

In this work, we propose to retain the tensor form of $\cX$ and assume that it admits the following low-rank decomposition,
\begin{equation}     
\label{eq:tucker}
\cX = \cS \times_1 \Ub_1 \times_2\cdots \times_d \Ub_d,
\end{equation}
where $\cS\in \mathbb{R}^{r_1\times \cdots\times r_d}$ is a core tensor, and $\Ub_1\in \mathbb R^{p_1\times r_1}, \ldots, \Ub_d\in \mathbb R^{p_d\times r_d}$ are matrices with orthonormal columns; see more details on this decomposition in Section \ref{sec:notation}. \change{We consider a low-rank model where the rank $r_i$ is much smaller than $p_i$. The low-rank assumption in \eqref{eq:tucker} exploits the structures in tensors and efficiently reduces the number of free parameters in $\cX$. Consider a special case where $r_i$ is fixed and $p_1=\ldots=p_d=p$ for simplicity. This low-rank modeling allows us to consider an efficient initialization phase with $\cO(p^{d/2})$ steps (see Lemma \ref{lemma:tensor_completion}), which is much reduced comparing to the $p^d$ steps required in the simple vectorization strategy. }
As demonstrate in Table \ref{table:com}, comparing to the vectorized solutions that ignore the low-rank structure, our proposed low-rank tensor bandit algorithms have much improved finite-time regret bounds.

Before discussing the main algorithms, we first describe our initialization procedure. 
Thanks to the tensor low-rank structure, our initialization phase need not to pull every arm at least once, which is required in the majority of multi-armed bandit algorithms. 
Define an initial set of $s_1$ steps
\begin{equation}\label{def:E_1}
\cE_1 = \{t \mid t\in[s_1]\},
\end{equation}
where $s_1$ is an integer to be specified later in Assumption \ref{con:initial}. 
In the initialization phase, arms are pulled with a uniform probability, equivalent to assuming $\mathbb P(i_{jt} = k) = 1/p_j$, $k\in[p_j]$, in \eqref{eqn:tensor_bandit}. 
If some prior knowledge about the true reward tensor is available, a non-uniform sampling can also be considered in the initialization phase.


\subsection{Tensor Elimination}\label{sec:tensor_elimination}

The upper confidence band (UCB) strategies \citep{lai1985asymptotically} have been very successful in bandit problems, leading to an extensive literature on UCB algorithms for standard multi-armed bandits \citep{lattimore2020bandit}. 
These UCB algorithms balance between exploration and exploitation based on a confidence bound that the algorithm assigns to each arm.
Specifically, in each round of steps, the UCB algorithm constructs an upper confidence bound for the reward associated with each arm, and the arms with the highest upper bounds are pulled, as they may be associated with high rewards and/or large uncertainties (i.e., under-explored). 
Many work have analyzed the regret bounds of UCB algorithms and investigated their optimality \citep{auer2002finite, garivier2011kl}. 

Employing the successful UCB strategy in low-rank tensor bandits encounters a critical challenge, as the tensor decomposition in \eqref{eq:tucker} is a non-convex problem, the data is adaptively collected and the concentration results for $\hat\cS, \hat{\Ub}_1,\ldots, \hat{\Ub}_d$, to our knowledge, remain elusive thus far. 
Without such concentration results, constructing the confidence bounds becomes a very difficult problem.
\change{One straightforward strategy is to first vectorize the tensor bandits and then treat the problem as a standard multi-armed bandit problem. However, as discussed before, this strategy incurs a severe loss of structural information and is demanding, in terms of sample complexity, in its initialization phase. 
In our proposed approach, we consider a tensor spectral-based rotation strategy that preserves the low-rank information and at the same time, enables uncertainty quantification. 
We also consider an elimination step that eliminate less promising arms based on the calculated confidence bounds, which further improves the finite-time regret bound (see Theorem \ref{thm:tensor_eliminator}).
Taken together, the proposed \texttt{tensor elimination} algorithm avoids directly characterizing the uncertainty of tensor decomposition estimators, effectively utilizes the low-rank information and achieves a desirable sub-linear finite-time regret bound.} 
Next, we discuss the \texttt{tensor elimination} algorithm in details.


\begin{algorithm}[htb!]
\caption{\texttt{Tensor elimination}}
\begin{algorithmic}[1]
\STATE \textbf{Input:} number of total steps $n$, number of exploration steps $n_1$, regularization parameters $\lambda_1,\lambda_2$, length of confidence intervals $\xi$, ranks $r_1,\cdots, r_d$.
\STATE { \# \emph{initialization and exploration phases}}
\STATE \textbf{Initialize:} $\cD = \emptyset$.
\FOR{$t= 1, \cdots, s_1+n_1$}
\STATE Randomly pull an arm $\cA_t$ and receive its associated reward $y_t$. Let $\cD = \cD \cup \{(y_t, \cA_t)\}$.
\ENDFOR
\STATE Calculate $\hat{\Ub}_1,\ldots,\hat{\Ub}_d$ using $\cD$, and then find $\hat{\Ub}_{1\perp},\ldots,\hat{\Ub}_{d\perp}$.
\STATE { \# \emph{reduction phase}}
\STATE Construct an action set $\mathbb A_1$ as in \eqref{eqn:linear_action_set} and denote $q = \Pi_{j=1}^d p_j-\Pi_{j=1}^d (p_j-r_j)$. \label{def:q}
\FOR{$k=1$ to  $ \log_2(n)$}
\STATE Set $V_{t_k} = \diag(\underbrace{\lambda_1, \ldots, \lambda_1}_{q}, \lambda_2, \ldots, \lambda_2)$ and  $\cD = \emptyset$.
\FOR{$t=t_k$ to $\min(t_{k+1}-1, n-n_1-s_1)$}
\STATE Pull the arm $A_t = \argmax_{a\in\mathbb A_k}\|\ab\|_{V_t^{-1}}$.
\STATE Receive its associated reward $y_t$ and update $V_{t+1} = V_t + A_t A_t^{\top}$. Let $\cD = \cD \cup \{(y_t, A_t)\}$.
\ENDFOR
\STATE Eliminate arms based on confidence intervals:
\begin{equation}\label{eqn:elimination}
\mathbb A_{k+1} = \Big\{\ab\in\mathbb A_k: \langle \hat{\bbeta}_k, \ab\rangle+\|\ab\|_{V_t^{-1}}\xi\geq \max_{\ab\in\mathbb A_k}\Big[\langle \hat{\bbeta}_k, \ab\rangle - \|\ab\|_{V_t^{-1}}\xi\Big]\Big\}, \text{ where }
\end{equation}
\begin{equation}\label{eqn:ridge}
\hat{\bbeta}_k =\argmin_{\bbeta}\Big\{\frac{1}{2}\sum_{(y_t, A_t)\in\cD}\big(y_t - \langle A_t, \bbeta\rangle\big)^2 + \frac{1}{2}\lambda_1\|\bbeta_{1:q}\|_{2} +\frac{1}{2} \lambda_2\|\bbeta_{(q+1):\Pi_{j=1}^d p_j}\|_{2}\Big\}.
\end{equation}
\ENDFOR
\end{algorithmic}
\label{alg:low_elim}
\end{algorithm}

The \texttt{tensor elimination} shown in Algorithm \ref{alg:low_elim} starts with an initialization phase of length $s_1$ and then proceeds to an exploration phase of length $n_1$, where arms in both phases are selected randomly. In this algorithm, the initialization phase and exploration phase are same. We choose to separate them so that the format is consistent with the \texttt{tensor epoch-greedy} algorithm introduced in next subsection.
Here, $s_1$ is set to be the minimal sample size for tensor completion and $n_1$ is chosen to minimize cumulative regret, both of which will be specified later in Section \ref{sec:tensor_elimination_theory}. \change{Based on the random samples collected from the initialization and exploration phases, we calculate estimates $\hat{\Ub}_1,\ldots, \hat{\Ub}_d$ of the matrices $\Ub_1,\ldots, \Ub_d$ in \eqref{eq:tucker} using a low-rank tensor completion method (see Appendix \ref{sec:tensor_alg}).
Next, we consider a rotation technique that preserves the tensor low-rank structure, and enables vectorization and uncertainty quantification (see Lemma \ref{lem:xi}).}
Specifically, given $\hat{\Ub}_j$, $j\in[d]$, define $\hat{\Ub}_{j\perp}$ whose columns are the orthogonal basis of the subspace complement to the column subspace of $\hat{\Ub}_j$. Consider a rotation to the true reward tensor $\cX$ calculated as  
\begin{equation*}
\cY = \cX \times_1 [\hat{\Ub}_1;\hat{\Ub}_{1\perp}]\times_2 \cdots \times_d[\hat{\Ub}_d;\hat{\Ub}_{d\perp}]\in\mathbb R^{p_1\times \cdots\times p_d},
\end{equation*}
where $\times_1,\ldots, \times_d$ are as defined in \eqref{eqn:marginal} and $[\hat{\Ub}_j;\hat{\Ub}_{j\perp}]$ is the concatenation (by columns) of $\hat{\Ub}_j$ and $\hat{\Ub}_{j\perp}$.
Correspondingly, the reward defined in \eqref{eqn:tensor_bandit} can be re-written (see proof in Appendix \ref{sec:equivalence}) as 
\begin{equation*}
y_t = \Big\langle \cY,[\hat{\Ub}_1;\hat{\Ub}_{1\perp}]^{\top}\eb_{i_{1,t}}\circ\cdots \circ [\hat{\Ub}_d;\hat{\Ub}_{d\perp}]^{\top}\eb_{i_{d,t}}\Big\rangle + \epsilon_t.
\end{equation*}
It is seen that replacing the reward tensor $\cX$ with $\cY$ and the arm $\eb_{i_1}\circ \cdots\circ \eb_{i_d}$ with $[\hat{\Ub}_1;\hat{\Ub}_{1\perp}]^{\top}\eb_{i_{1,t}}\circ\cdots \circ [\hat{\Ub}_d;\hat{\Ub}_{d\perp}]^{\top}\eb_{i_{d,t}}$ does not change the tensor bandit problem. 
Define $\bbeta=\text{vec}(\cY)\in \mathbb R^{\prod_{j=1}^d p_j}$, which vectorizes the reward tensor $\cY$ such that the first $\Pi_{j=1}^dr_j$ entries of $\text{vec}(\cY)$ are $\cY_{i_1,\ldots, i_d}$ for $i_j\in\{1,\ldots, r_j\}$, $j\in[d]$, and denote the corresponding vectorized arm set as
\begin{eqnarray}\label{eqn:linear_action_set}
\mathbb A := \Big\{\text{vec}\Big([\hat{\Ub}_1;\hat{\Ub}_{1\perp}]^{\top}\eb_{i_1}\circ \cdots \circ[\hat{\Ub}_d;\hat{\Ub}_{d\perp}]^{\top}\eb_{i_d} \Big), i_1\in[p_1], \ldots, i_d\in [p_d]\Big\}.
\end{eqnarray}
Correspondingly, the tensor bandits in \eqref{eqn:tensor_bandit} with the true reward tensor $\cX$ and arm set $\left\{\eb_{i_1}\circ \cdots \circ \eb_{i_d}, i_1\in[p_1],\cdots,i_d\in [p_d]\right\}$ can be re-formulated as a multi-armed bandits with the reward vector $\bbeta$ and arm set $\mathbb A$. 

It is easy to see that in $\text{vec}(\cX \times_1 [{\Ub}_1;{\Ub}_{1\perp}]\times_2 \cdots \times_d[{\Ub}_d;{\Ub}_{d\perp}])$, the first $\Pi_{j=1}^dr_j$ entries are nonzero and the last $\Pi_{j=1}^d(p_j-r_j)$ entries are zero. Such a sparsity pattern cannot be achieved if $\cX$ is vectorized directly without the rotation. From this perspective, the rotation strategy preserves the structural information in the vectorized tensor. 
Specifically, when estimating the reward vector $\bbeta$ in \eqref{eqn:ridge}, we apply different regularizations to the first $\Pi_{j=1}^dr_j$ entries and the remaining $\Pi_{j=1}^d(p_j-r_j)$ entries, respectively.


The algorithm then proceeds to the elimination phase, where less promising arms are identified and eliminated. This phase aims to further improve the regret bound. 
Given a vector $\ab$, we define its $\Ab$-norm as $\|\ab\|_\Ab = \sqrt{\ab^\top \Ab \ab}$, where $\Ab$ is a positive definite matrix.
During phase $k$ with the arm set $\mathbb{A}_k$, the confidence ellipsoid of the mean reward of each arm $\ab\in\mathbb{A}_k$ is constructed using $\hat{\bbeta}_k$. 
It is shown in Lemma \ref{lem:xi} that the confidence width of the reward of arm $\ab$ is $\|\ab\|_{V^{-1}}\xi$, where $V$ is the covariance matrix and $\xi$ is a fixed constant term that does not depend on $\ab$. 
At each time step $t$, the algorithm (line 13) then pulls the arm with the largest confidence interval width. 
The intuition of the arm selection in this step is that arms with the highest confidence widths are likely under-explored. 
At the end of phase $k$ (line 16), we implement an elimination procedure that trims less promising arms.
Specifically, we first update the estimate $\hat{\bbeta}_k$ in \eqref{eqn:ridge} based on the pulled arms and their associated rewards during phase $k$. 
Based on the estimated reward $\hat{\bbeta}_k$,  we then construct confidence interval \eqref{eqn:elimination} for the mean reward of each arm and eliminate the arms whose upper confidence bound is lower than the maximum of lower confidence bounds of all arms in $\mathbb{A}_k$.

\subsection{Regret Analysis of Tensor Elimination}\label{sec:tensor_elimination_theory}
In this section, we carry out the regret analysis of the \texttt{tensor elimination}. 
To ease notation, we assume the tensor rank $r_1=\ldots=r_d=r$ and the tensor dimension $p_1=\ldots=p_d=p$. 
The results for general ranks and dimensions can be established similarly using a more involved notation system. We first state some assumptions.
 
\begin{assumption}[Sub-Gaussian noise]
\label{con:noise}
The noise term $\epsilon_t$ is assumed to follow a 1-sub-Gaussian distribution such that, for any $\lambda\in\mathbb R$,
\begin{equation*}
\mathbb E[\exp(\lambda\epsilon_t)]\leq \exp(\lambda^2/2).
\end{equation*}
\end{assumption}
\noindent

\begin{assumption}
\label{con:max}
Assume true reward tensor $\cX$ admits the low-rank decomposition in \eqref{eq:tucker} with $r = \cO(p^{1/(d-1)})$. In addition, we assume $\|\cX\|_{\infty}\leq 1$ and \change{$p^{d/2}\|\cX\|_{\infty}/\|\cX\|_{F}=\cO(1)$}.
\end{assumption}
\noindent
The assumption $\|\cX\|_{\infty}\leq 1$ assumes that the reward is bounded, and it is common in the multi-armed bandit literature \citep[see, for example,][]{langford2007epoch}. 
It implies that the immediate regret in each exploration step is $\cO(1)$. 
Similar boundedness conditions on tensor entries can also be found in the tensor completion literature \cite[see, for example,][]{cai2021nonconvex,xia2021statistically}. The assumption on the rank $r$ refers to a low-rank model assumption and is to simplify the final sample size requirement. \change{Moreover, $p^{d/2}\|\cX\|_{\infty}/\|\cX\|_{F}$ measures the spikiness of the true tensor and its boundedness ensures that low-rank tensor completion based on randomly observed samples can be reliable. This condition is a typical incoherence assumption that is used in \cite{xia2021statistically,cai2021nonconvex} and is also common in other low-rank models \citep{negahban2012restricted}. 
}

\begin{assumption}
\label{con:initial}
Assume the number of steps in the initialization phase $s_1$ is
\begin{eqnarray}\label{eqn:s_1}
s_1 = C_0 r^{(d-2)/2}p^{d/2},
\end{eqnarray}
where $C_0$ is a positive constant as defined in Lemma \ref{lemma:tensor_completion}.
\end{assumption}
\noindent

This assumption requires the minimal sample complexity for provably recovering a low-rank tensor from noisy observations when the entries are observed randomly (see Lemma \ref{lemma:tensor_completion} and \cite{xia2021statistically}). Such random initialization phase is standard and important in all bandit algorithms \citep{lattimore2020bandit}. As discussed before, the simple vectorization strategy would require $s_1=\cO(p^d)$, which is significantly larger. 

The next lemma provides the confidence interval for the reward of a fixed arm $\ab$. 
\begin{lemma}\label{lem:xi}
For any fixed vector $\ab\in\mathbb{R}^{p^d}$ and $\delta>0$, we have that, if 
\begin{equation}\label{eqn:xi}
\xi= 2\sqrt{14\log(2/\delta)}+\sqrt{\lambda_1} \|\bbeta_{1:q}\|_{2} + \sqrt{\lambda_2}\|\bbeta_{(q+1):p^d}\|_{2},
\end{equation}
with $\bbeta = vec(\cY)$, $\lambda_1>0$ and $\lambda_{2} = n/(q\log(1+n/\lambda_1))$, then at the beginning of phase $k$ 
\begin{equation*}
\mathbb{P}(|\ab^\top(\hat{\bbeta}_k-\bbeta) | \le \xi\|\ab\|_{V_t^{-1}})\ge 1-\delta,
\end{equation*}
where $V_t = \sum_{s=1}^{t}A_sA_s^{\top}+ \diag(\underbrace{\lambda_1, \ldots, \lambda_1}_{q}, \lambda_2, \ldots, \lambda_2)$.
\end{lemma}

Next, we show the finite-time regret bound for \texttt{tensor elimination}. Recall $q = \Pi_{j=1}^d p_j-\Pi_{j=1}^d (p_j-r_j)$.
\begin{theorem}\label{thm:tensor_eliminator}
Suppose Assumptions \ref{con:noise}-\ref{con:initial} hold. Let $t_k =2^{k-1}$, $0<\lambda_1 \leq1/p^d$, $\lambda_{2} = n/(q\log(1+n/\lambda_1))$, and
\begin{equation}\label{eqn:n_1}
n_1 = \left\lceil n^{^{\tfrac{2}{d+2}}}\frac{r^{d}}{\Pi_{j=1}^{d}\sigma_j}p^{\frac{d^2+d}{2}}\log^{d/2} (p) \right\rceil,
\end{equation}
where $\sigma_j$ is the 
smallest non-zero singular value of $\cM_j(\cX)$, $j\in[d]$. 
The cumulative regret of Algorithm \ref{alg:low_elim} satisfies
\begin{equation*}
\begin{split}
\textrm{R}_n\leq C\Big(r^{\tfrac{d}{2}}p^{\tfrac{d}{2}} +\Big(\frac{r^{d}}{\Pi_{i=1}^d\sigma_i}\log^{d/2} (p)\Big)^{\tfrac{2}{d+2}}p^{\tfrac{d^2+d}{d+2}}n^{\tfrac{2}{d+2}} + \sqrt{(d\log(p)+\log(n))^2p^{d-1}n}\Big),
\end{split}
\end{equation*}
with probability at least $1-dp^{-10}-1/n$, where $C > 0$ is some constant. 
\end{theorem}

\noindent
The detailed proof of Theorem \ref{thm:tensor_eliminator} is deferred to Appendix \ref{proof:tensor_eliminator}. \change{It should be noted that this paper focuses on the high-dimensional setting, where $p$ approaches infinity, to ensure the probability approaches 1. A similar prerequisite of $p\rightarrow \infty$ is also essential in both the bilinear matrix bandits \citep{jun2019bilinear} and low-rank tensor model \citep{xia2021statistically} to ensure that the probability approaches 1.probability approaches 1.}
Ignoring any logarithmic and constant factor, the above regret bound can be simplified to 
\begin{equation}\label{eqn:elim_bound}
\textrm{R}_n=\tilde{\cO}(r^{\tfrac{d-2}{2}}p^{\tfrac{d}{2}} + r^{\tfrac{2d}{d+2}}p^{\tfrac{d^2+d}{d+2}}n^{\tfrac{2}{d+2}} + p^{\tfrac{d-1}{2}}n^{\tfrac{1}{2}}).
\end{equation}
The upper bound on the cumulative regret is the sum of three terms, with the first two terms characterizing regret from the $s_1$ initialization steps and $n_1$ exploration steps, respectively, and the third term quantifying the regret in the $n-s_1-n_1$ elimination steps. 
As the regret from the exploration phase increases with $n_1$ and the regret from the elimination phase decreases with $n_1$, the value for $n_1$ in \eqref{eqn:n_1} is chosen to minimize the sum of these two regrets. Note that after the rotation, the order of $\|\bbeta_{1:q}\|_2$ is of $\tilde{\cO}(p^{d/2})$ which guides the choice of $\lambda_1$.
One component of the upper bound of the cumulative regret is $\log\left(\frac{\det(V_k)}{\det(\Lambda)}\right)$ with  $\Lambda = \diag(\lambda_1,\ldots,\lambda_1, \lambda_2, \ldots, \lambda_2)$ 
and $\lambda_2$ is chosen to minimize the upper bound of the log term so as to minimize the upper bound of cumulative regret. 
\change{Theorem \ref{thm:tensor_eliminator} is derived assuming $r$ is considerably smaller than $p$. In the case of a full-rank tensor with $r=p$, there is no benefit of considering a low-rank model and one could unfold the tensor into a long vector and employ an existing bandit algorithm such as LinUCB, which has been shown to be optimal in linear bandits.}

\begin{remark}
It is worth to compare the regret bound in Eq.~\eqref{eqn:elim_bound} with other strategies. As summaized in Table \ref{table:com}, when $d= 3$ and $r = \cO(1)$, \texttt{vectorized UCB} suffers $\tilde{\cO}(p^3+p^{3/2}n^{1/2})$. If we unfold the tensor into a matrix and implement ESTR \citep{jun2019bilinear}, it suffers $\tilde{\cO}(p^2+p^3n^{1/2})$. Both of these competitive methods obtain significantly sub-optimal regret bounds. By utilizing the low-rank tensor information, our bound greatly improves the dependency on the dimension $p$. Moreover, our advantage is even larger when the tensor order $d$ is larger. 
\end{remark}

\begin{remark}
One may wonder whether we can extend the matrix bandit ESTR \citep{jun2019bilinear} to the tensor case. In this case, standard LinUCB \citep{abbasi2011improved} algorithm could be queried to handle the reshaped linear bandits as did in the matrix bandits \citep{jun2019bilinear}. However, it is known that the algorithm of LinUCB is suboptimal for linear bandits with finitely many arms and the sub-optimality will be amplified as the order of tensor grows. Hence, using LinUCB in the reduction phase results in $\cO(p^2n^{1/2})$ for the leading term that is even worse than \texttt{vectorized UCB}.
\end{remark}


One of the key challenges in our theoretical analysis is to quantify the cumulative regret in the elimination phase. Existing techniques are not applicable as we utilize a different eliminator with a modified regularization strategy. Furthermore, to bound the cumulative regret in the elimination phase, we need to bound the norm $\|\bbeta_{(q+1):p^d}\|_2$ which is the last $p^d-q$ entries of $vec(\cY)$. Recall that the reward tensor $\cY$ is a rotation of true reward tensor $\cX$. 
We need to derive the upper bound of the norm of rotated reward vector by exploiting the knowledge of estimation error of $\cX$. We use the elliptical potential lemma to bound the cumulative regret in elimination phase. Furthermore, all parameters such as the penalization parameter $\lambda_2$ and exploration phase length $n_1$ are carefully selected to obtain the best bound. 


\subsection{Tensor Epoch-greedy and Regret Analysis}
Next, we propose an epoch-greedy type algorithm for low-rank tensor bandits, and compare its performance with \texttt{tensor elimination}.
The epoch-greedy algorithm \citep{langford2007epoch}
proceeds in epochs, with each epoch consisting of an exploration phase and an exploitation phase. One advantage of this epoch-greedy algorithm is that we do not need to know the total time horizon $n$ in advance. 
In the exploration phase, arms are randomly selected and in the exploitation phase, arms that expect the highest reward are pulled.
The number of steps in each exploitation phase increases with number of epochs, guided by the fact that, as the number of epochs increases, the estimation accuracy of the true reward improves and more exploitation steps are desirable.
The epoch-greedy algorithm is straightforward to implement, and we find that compared to \texttt{tensor elimination}, \texttt{tensor epoch-greedy} algorithm has a better dependence on dimension $p$ and a worse dependence on time horizon $n$. 

\begin{algorithm}[t!]
\caption{ \texttt{Tensor epoch-greedy}}
\begin{algorithmic}[1]\label{alg:epoch_greedy}
\STATE \textbf{Input:} initial set $\cE_1$, exploration set $\cE_2$.
\STATE Initialize $\cD = \emptyset$.
\FOR{$t=1,2,\ldots, n$}
\STATE \# \emph{initialization and exploration phases}
\IF{$t\in \cE_1\cup \cE_2$}
\STATE Randomly pull an arm $\cA_t$ and receive its associated reward $y_t = \langle \cX, \cA_t\rangle + \epsilon_t.$
\STATE  Let $\cD = \cD \cup \{(y_t, \cA_t)\}$.
\ENDIF
\STATE \# \emph{exploitation phase}
\IF{$t\notin\cE_1\cup \cE_2$}
\STATE Based on $\cD$, calculate a low-rank tensor estimate $\hat{\cX}_t$.
\STATE Pull the arm $(i_{1,t}, \ldots, i_{d,t}) = \argmax_{i_{1},\ldots, i_{d}}\langle \hat{\cX}_t, \eb_{i_{1}}\circ \ldots \circ\eb_{i_{d}} \rangle.$
\STATE Receive the associated reward $y_t = \langle \cX,  \eb_{i_{1,t}}\circ \ldots \circ \eb_{i_{d,t}}\rangle + \epsilon_t.$
\ENDIF
\ENDFOR
\end{algorithmic}
\end{algorithm}
	
The detailed steps of \texttt{tensor epoch-greedy} are given in Algorithm \ref{alg:epoch_greedy}. 
In the initialization phase, i.e., $t\in \cE_1$, arms are randomly pulled to collect samples for tensor completion. Recall the initialization phase has $s_1$ steps.
Let the index set of steps in the exploration phases be
\begin{equation}\label{def:explor}
\cE_2 = \Big\{s_1+ l+1+\sum_{k=0}^ls_{2k} \ | \ l=0,1,\ldots\Big\},
\end{equation}
where $s_{2k}$ denotes the number of exploitation steps in the $k$th epoch and it increases with $k$. 
In the exploration phase, i.e., $t\in\cE_2$, an arm $\cA_t$ is pulled (or sampled) randomly. 
These random samples collected in the exploration phases are important for unbiased estimation, as they do not depend on historical data, and their accumulation can improve estimation accuracy of the reward tensor.
Meanwhile, as the exploration phase does not focus on the best arm, each step $t\in\cE_2$ is expected to result in a large immediate regret, though it can potentially reduce regret from future exploitation steps. 
In the exploitation phase, i.e., $t\notin\cE_1\cup \cE_2$, we construct a low-rank estimate $\hat{\cX}_t$ of the reward tensor using the random samples collected thus far in $\cD$.
Then, the arm $(i_{1,t}, \ldots, i_{d,t})$ with the highest estimated reward in $\hat{\cX}_t$ is selected, i.e.,
\begin{equation*}
(i_{1,t}, \ldots, i_{d,t}) = \argmax_{i_{1},\ldots, i_{d}}\big\langle\hat{\cX}_t, \eb_{i_{1}}\circ \ldots \circ \eb_{i_{d}} \big\rangle.
\end{equation*}
Samples in the exploitation phase will not be used to estimate the reward tensor as they are biased and thus exploitation steps cannot improve estimation accuracy of the reward tensor. 


We next derive the regret bound of proposed \texttt{tensor epoch-greedy}. 
\begin{theorem}\label{thm:tensor_etc}
Suppose Assumptions \ref{con:noise}-\ref{con:initial} hold. Let 
\begin{equation}\label{def:number_exploration}
s_{2k}= \left\lceil C_2 p^{-\tfrac{d+1}{2}}r^{-\tfrac{1}{2}}(\log p)^{-\tfrac{1}{2}}(k+s_1)^{\tfrac{1}{2}} \right\rceil,
\end{equation}
for some small constant $C_2>0$. 
When $n\geq C_0r^{\frac{d-2}{2}}p^{\frac{d}{2}}$, the cumulative regret of Algorithm \ref{alg:epoch_greedy} satisfy, with probability at least $1-p^{-10}$,
\begin{equation}\label{eqn:regret_tensor_bandit}
\textrm{R}_n\leq C_0r^{\frac{d-2}{2}}p^{\frac{d}{2}} + 8n^{\tfrac{2}{3}}p^{\tfrac{d+1}{3}}(r\log p)^{\tfrac{1}{3}}.
\end{equation}
\end{theorem}

\noindent
The regret bound has two terms with the first term characterizing the regret accumulated during the initialization phase and the second term characterizing the regret accumulated over the exploration and exploitation phases.
The first term depends on the tensor rank $r$ and dimension $p$, but not $n$. It clearly highlights the benefit of exploiting a tensor low-rank structure since unfolding the tensor into a vector or a matrix requires much longer initialization phase. The second term in the regret bound is related to time horizon $n$ and it increases with $n$ at a rate of $n^{\tfrac{2}{3}}$. 


It is worth to compare the leading term of  regret bounds for high-order tensor bandits of \texttt{tensor elimination} in Eq.~\eqref{eqn:elim_bound} and \texttt{tensor epoch-greedy} in Eq.~\eqref{eqn:regret_tensor_bandit}. As summaized in Table \ref{table:com}, when $d\geq 3$ and $r = \cO(1)$, \texttt{tensor elimination} suffers $\tilde{\cO}(p^{(d-1)/2}\sqrt{n})$ regret while \texttt{tensor epoch-greedy} suffers  $\tilde{\cO}(p^{(d+1)/3}n^{2/3})$ regret. Although the latter one has a sub-optimal dependency on the horizon due to the $\varepsilon$-greedy paradigm, it enjoys a better regret than the prior one in the high-dimensional regime ($n\leq p^{d-5}$).

In the theoretical analysis, a key step is to determine the switch time between the two phases, i.e., $s_{2k}$. We set the length of exploitation phase to be the inverse of tensor estimation error. Intuitively, when the tensor estimation error is large, more exploration can increase the sample size and improve the estimation. When the tensor estimation error is small, there is no need to perform more randomly exploration. Instead, we exploit more to reduce instant regrets. After obtaining the regret in epoch, we need to derive the upper bound of number of epochs. Similar to the optimal tuning procedure in explore-then-commit regret analysis, we tune the parameter to determine the final bound of total number of exploration steps.


\section{Contextual Tensor Bandits}
\label{sec:contextual_bandit}


In this section, we consider an extension of tensor bandits to contextual tensor bandits where some modes of the reward tensor are contextual information. Take the online advertising data considered in Section \ref{sec:real} as an example. Users use the online platform on some day of the week, and the platform can only decide which advertisement to show to this given user at the given time. In this example, the user mode and the day-of-week mode of the reward tensor are both contextual information and both are not decided by the platform.



The above example can be formalized as contextual tensor bandits. 
Specifically, at time $t$, the agent observes a $d_0$-dimensional context $(i_{1,t},\cdots, i_{d_0,t})\in [p_1]\times\cdots\times [p_{d_0}]$ and given the observed context, pulls an $(d-d_0)$-dimensional arm $(i_{d_0+1,t},\cdots, i_{d,t})\in [p_{d_0+1}]\times\cdots\times [p_{d}]$. 
Let $I_t =(i_{1,t},\cdots, i_{d,t})$ collect the \textit{context}$\times$\textit{arm} information at time step $t$. 
Correspondingly, the agent observes a noisy reward $y_t$ drawn from a probability distribution associated with $I_t$. The objective is to maximize the cumulative reward over the time horizon. 
This contextual tensor bandit problem is different from the tensor bandit problems considered in Section \ref{sec:sec3}, as 
the agent does not have the ability to choose the context. 
Therefore, the \texttt{tensor elimination} 
algorithm can not be applied to contextual tensor bandits. 
To tackle this problem, we introduce a heuristic solution to contextual tensor bandits that utilizes Thompson sampling \citep{russo2018tutorial} and ensemble sampling \citep{lu2017}.

Thompson sampling is a powerful Bayesian algorithm that can be used to address a wide range of online decision problems. The algorithm, in its basic form, first initializes a prior distribution over model parameters, and then samples from its posterior distribution calculated using past observations. Finally, an action is made to maximize the reward given the sampled parameters. 
The posterior distribution can be derived in closed-form in a few special cases such as the Bernoulli bandit \citep{russo2018tutorial}. 
With more complex models such as our low-rank tensor bandit problem, the exact calculation of the posterior distribution may become intractable. In this case, we consider an ensemble sampling approach that aims to approximate Thompson sampling while maintaining computational tractability. Specifically, ensemble sampling aims to maintain, incrementally update, and sample from a finite ensemble of models; and this ensemble of models approximates the posterior distribution \citep{lu2017}.

Consider the true reward tensor $\cX\in\mathbb R^{p_1\times \ldots\times p_d}$ that admits the decomposition in \eqref{eq:tucker}, where the first $d_0$ dimensions of $\cX$ correspond to the context and the last $d-d_0$ dimensions correspond to the decision (or arm). 
At time $t$ and given the arm $\cA_t = \eb_{i_{1,t}}\circ\cdots \circ \eb_{i_{d,t}}$, the reward $y_t$ is assumed to follow $y_t = \langle\cX,\cA_t\rangle+\epsilon_t$. To ease the calculation of the posterior distribution, in contextual tensor bandits we consider $\epsilon_t\sim N(0,\sigma^2)$. For the prior distribution over model parameters, we assume the rows of $\Ub_k$ are drawn independently from 
\begin{equation*}
    [\Ub_k]_{i,\cdot}\sim N(\bm{\mu}_{k,i},\sigma_k^2\Ib), \quad i\in[p_k],\,\,k\in[d].
\end{equation*}
Let $\cH_{t-1}=\{(\cA_{{s}},y_{{s}})\}^{t-1}_{{s} =1}$ denote the history of action$\times$reward up to time $t$. Given the prior distribution, the posterior density function can be calculated as 
$$
f(\cX|y_1,\cdots, y_{t-1})\propto f(y_1\cdots, y_{t-1}|\cX)\Pi_{k,i}f([\Ub_k]_{i,\cdot}).
$$
 We maximize $f(\cX|y_1,\cdots, y_{t-1})$ to obtain the maximum the posteriori (MAP) estimate as
\begin{eqnarray}
(\hat{\cS}^{(t)}, \hat{\Ub}_1^{(t)},\cdots,\hat{\Ub}^{(t)}_d) = \argmin_{\cS, \Ub_{1},\cdots,\Ub_d}\Big(\frac{1}{\sigma^2}\sum_{{s} =1}^{t-1}(y_{{s}}-\langle\cX,\cA_{s}\rangle)^2+\sum_{k=1}^{d}\frac{1}{\sigma_k^2}\sum_{i=1}^{p_k}\Big\|[\Ub_k]_{i,\cdot}-\bm\mu_{k,i}\Big\|_2^2\Big).
\label{eqn:objective}
\end{eqnarray}
The objective function in \eqref{eqn:objective} can be equivalently written as 
\begin{eqnarray*}
\frac{1}{\sigma^2}\sum_{{s} =1}^{t-1}(y_{{s}}-\cS\times_1 [\Ub_1]_{i_{1{s}},\cdot}\times \cdots \times_d [\Ub_d]_{i_{d{s}},\cdot})^2+\sum_{k=1}^{d}\frac{1}{\sigma_k^2}\sum_{i=1}^{p_k}\Big\|[\Ub_k]_{i,\cdot}-\bm\mu_{ki}\Big\|_2^2,
\end{eqnarray*}
which is a non-convex optimization problem. 
In our proposed algorithm, we alternatively optimize $\Ub_{k}$, $k\in[d]$ and $\cS$. Given all $\Ub_{l}$ such that $l\neq k$ and $\cS$, we estimate the $i$-th row of $\Ub_{k}$ as
\begin{eqnarray*}
	[\Ub_{k}]_{i,\cdot}^{(t)}&=&\bigg[\frac{1}{\sigma^2}\sum_{{s}=1}^{t-1}\mathbf{1}_{(i_{k{s}}=i)}\vb^{(t-1)}(\vb^{(t-1)})^{\top}+\frac{1}{\sigma_{1}^2}\Ib\bigg]^{-1}\bigg\{\frac{1}{\sigma^2}\sum_{{s}=1}^{t-1}\mathbf{1}_{(i_{1,{s}}=i)}y_{{s}}\vb^{(t-1)}+\frac{1}{\sigma^2}\bm\mu_{k,i}\bigg\},
\end{eqnarray*}
where $\vb^{(t-1)} = \left\{\cS^{(t-1)}\times_1[\Ub_1]^{(t-1)}_{i_{1{s}},\cdot}\times \cdots\times_{k-1}[\Ub_{k-1}]^{(t-1)}_{i_{k-1,{s}},\cdot}\times _{k+1}[\Ub_{k+1}]^{(t-1)}_{i_{k+1,{s}},\cdot}\times \cdots\times_d [\Ub_d]^{(t-1)}_{i_{d{s}},\cdot}\right\}$. After updating all rows of $\Ub_k$ for $k\in[d]$, we then estimate $\cS$ by solving \eqref{eqn:objective}.

\begin{algorithm}[t!]
\caption{\texttt{Tensor ensemble sampling}}
\begin{algorithmic}[1]
\STATE \textbf{Input:} rank $r_1,\ldots,r_d$, $\sigma^2$, $\{\bm\mu_{ki}\}_{i\in[p_k], k\in[d]}$, $\{\sigma^2_k\}_{k\in[d]}$, number of models $M$, variance of perturbed noise $\tilde{\sigma}^2$.
\STATE { \# \emph{initialize $M$ models from prior distributions }}
\STATE \textbf{Initialize} sample $[\hat{\Ub}_{km}]^{(0)}_{i,\cdot}\sim N(\bm\mu_{ki},\sigma^2_k\Ib)$ for $m\in[M]$, $i\in[p_k]$, $k\in[d]$. Normalize each column of matrix $\hat{\Ub}_{km}^{(0)}$. Initialize the core tensor $\cS_{m}^{(0)}=\mathbf{1}\circ\cdots\circ\mathbf{1}\in \mathbb{R}^{r_1\times_1\cdots\times_{d}r_d}$.
\STATE \textbf{for} t = $1,2\cdots$ \textbf{do}
\STATE { \# \emph{exploitation phase}}
\STATE \hspace{0.15in} Sample $\tilde{m}\sim \text{Unif}\{1,\cdots,M\}$
\STATE \hspace{0.15in} Observe context $\bx_t=(i_{1t},\cdots,i_{d_0t})$
\STATE \hspace{0.15in} Update $(\hat{\cS}_{\tilde{m}}^{(t)},\hat{\Ub}^{(t)}_{1\tilde{m}},\cdots,\hat{\Ub}^{(t)}_{d\tilde{m}})$ by solving \eqref{eqn:perturb}.\label{lst:line:max}
\STATE \hspace{0.15in} Choose $\ab_t=(i_{d_0+1,t},\cdots,i_{dt})=\argmax_{\substack{\ab=(i_{d_0+1},\cdots,i_{d})\in [p_{d_0+1}]\times \cdots\times [p_d]}}\hat{R}_{\tilde{m}}(\bx_t,\ab)$, where 
$$\hat{R}_{\tilde{m}}(\bx_t,\ab) = \hat{\cS}_{\tilde{m}}^{(t)}\times_1 [\hat{\Ub}_{1\tilde{m}}]^{(t)}_{i_{1t},\cdot}\times \cdots\times_{d} [\hat{\Ub}^{(t)}_{d\tilde{m}}]_{ i_{dt},\cdot}.$$
\STATE \hspace{0.15in} Receive reward $y_t$.	
\STATE { \# \emph{perturbation phase}}
\STATE \hspace{0.15in} Sample perturbation  noise $\omega_{tm}\sim N(0,\tilde{\sigma}^2)$ for $m\in[M]$.
\STATE \hspace{0.15in} Obtain perturbed rewards $\tilde{y}_{tm}=y_t+\omega_{tm}$ for $m\in[M]$.
\STATE \textbf{end} \textbf{for} 		
\end{algorithmic}
\label{alg:ensemble}
\end{algorithm}

\texttt{Tensor ensemble sampling} in Algorithm \ref{alg:ensemble} consists of initialization, exploitation and perturbation phases. 
In the initialization phase, we sample $M$ models from the prior distributions. The mean $\bm\mu_{ki}$ and variance $\sigma_k^2$ in the prior distributions could be determined from prior knowledge or specified so that the range of models spans plausible outcomes. 
Then, at each time step $t$, a model $\tilde{m}$ is uniformly  sampled from the ensemble of $M$ models. After observing a context $\bx_t = (i_{1t},\cdots,i_{d_0t})$, the agent exploits the history data of model $\tilde{m}$ to estimate the low-rank component of the reward tensor via
\begin{equation}
\min_{\cS,\Ub_1,\cdots,\Ub_d}\frac{1}{\sigma^2}\sum_{{s} =1}^{t-1}\Big\{\tilde{y}_{{s} m}-\langle\cX,\cA_{{s}}\rangle\Big\}^2+\sum_{k=1}^{d}\frac{1}{\sigma_k^2}\sum_{i=1}^{p_k}\Big\|[\Ub_k]_{i,\cdot}-[\hat{\Ub}_{k,m}]^{(0)}_{i,\cdot}\Big\|_2^2.
\label{eqn:perturb}
\end{equation}
Compared to \eqref{eqn:objective}, the objective in \eqref{eqn:perturb} uses perturbed rewards and perturbed priors, which helps to diversify the models and capture model uncertainty. The goal is for the ensemble to approximate the posterior distribution and the variance among models to diminish as the posterior concentrates. 
Based on the sampled model $\tilde{m}$, we pull the optimal arm $\ab_t$ given the observed context $\bx_t$. At the end of each time step, we perturb observed rewards for all $M$ models to diversify the ensemble. 
Our \texttt{tensor ensemble sampling} can be viewed as an extension of ensemble sampling \citep{lu2017} for contextual bandits problem. \change{Note that \eqref{eqn:perturb}  is a non-convex optimization problem, and there is no assurance of achieving the global optimizer. However, the optimization problem in \eqref{eqn:perturb} is bi-convex, meaning that the loss function is convex with respect to one set of parameters while fixing the other sets.  
This attractive property guarantees that the algorithm will always converge, though possibly to a local optimum \citep{xu2013block}. \change{Whether the algorithm can reach the global optimum depends on how close the initialization value is to the true value.} 
The same holds for other similar low-rank tensor estimation problems \citep{Sun2016, cai2021nonconvex, xia2021statistically}. 
In all of our experiments, we have observed that the \texttt{tensor ensemble sampling} method performs 
well with the random initialization utilized in Algorithm 3. It is challenging to analytically quantify how local solutions to \eqref{eqn:perturb}  affect the \texttt{tensor ensemble sampling} method and we leave a comprehensive theoretical investigation to future work.
Moreover, our choices of Gaussian prior distribution and Gaussian perturbation noise follow from the existing ensemble sampling literature \citep{lu2017, osband2018randomized, kveton2020randomized, dwaracherla2022ensembles, NEURIPS2022_874f5e53} due to their successful empirical performance and ease in computation in practice.}

\change{
Although \texttt{tensor ensemble sampling} is motivated by contextual tensor bandit problems, it can also be used to solve tensor bandits without context. In this case, the context dimension $d_0=0$ and an arm $\cA_t$ consists of all decisions to be made. 
While \texttt{tensor ensemble sampling} performs well empirically, its theoretical investigation 
is very challenging due to the 
nature of the ensemble sampling framework \citep{lu2017} and the non-convex optimization in low-rank tensor problems. In Section \ref{app:ats} of the supplement, we present some preliminary Bayes regret analysis of a general approximate Thompson sampling (TS) algorithm 
for tensor bandits. Notably, our \texttt{tensor ensemble sampling} can be considered as a specific instance of an approximate TS algorithm. Since approximate TS is a Bayesian algorithm, following the literature in this field \citep{russo2016information, NEURIPS2022_874f5e53}, we develop a Bayes regret bound, rather than a frequentist regret bound in $(\ref{eqn:pseduo_regret})$. 
The Bayes regret is defined as
$$
\textrm{BR}_n = \mathbb E \left [ \textrm{R}_n \right ] = \mathbb E \left[ \sum_{t=1}^n \left \langle \cX, \cA^* \right \rangle - 
\sum_{t=1}^n \left \langle \cX, \cA_t \right \rangle
\right],
$$
where the expectation is taken over the reward tensor $\cX$ under the prior distribution $P_0$. Different from the frequentist regret bound in $(\ref{eqn:pseduo_regret})$, the Bayes regret has an additional expectation over the reward tensor $\cX$. Theorem \ref{thm:bayes-regret} in Section \ref{app:ats-regret} provides a Bayes regret bound for a general approximate TS, 
$
\textrm{BR}_n \leq \mathcal{\tilde{O}} \left( \sqrt{p^d \, \mathbb{H}(A^*) n} + \sum_{t=1}^n \mathbb{E} \left[\mathbf{d}_{\mathrm{H}} (P_t^* \| \bar{P}_t)\right] \right),
$
where $\mathbb{H}(A^*)$ is the entropy of optimal action $A^*$ and the second term measures the distance between its action sampling distribution $\bar{P}_t(\cdot)= {\tt sample}(\cdot \, | \, P_0, \cD_{t-1})$, and that of the standard Thompson sampling algorithm, $P^*_t (\cdot)=\Pr(A^* \in \cdot | \cD_{t-1})$. See Section \ref{app:ats-regret} for more details. 


It is important to note that the above Bayesian regret bound is based on a preliminary information-theoretic analysis and we expect its dependence on $p$, the dimension of the tensor, can be further improved. Specifically, our analysis has not fully exploited the low-rank structure of the reward tensor $\cX$. The question of how to incorporate this low-rank structure into the information-theoretic analysis of approximate Thompson sampling remains an open problem that is particularly challenging. We believe that addressing this problem will necessitate novel insights into Bayesian inference in low-rank tensors and potentially require additional assumptions about the prior distribution $P_0$. Even in the low-rank matrix case, this issue is not well understood, and we see it as an interesting but very challenging direction for future research. Moreover, to derive an explicit regret bound for the \texttt{tensor ensemble sampling} algorithm, we need to further bound the Hellinger distance term $\mathbf{d}_{\mathrm{H}} (P_t^* \| \bar{P}_t)$ for the ensemble sampling algorithm. We believe that this is also a challenging problem that requires better understanding of how perturbed rewards and priors affect the non-convex tensor decomposition formulation (see equation~(\ref{eqn:perturb})), as well as their connections to Bayesian inferences in low-rank tensors. It is worth mentioning that \citet{NEURIPS2022_874f5e53} has provided a Bayes regret bound for ensemble sampling in a special linear Gaussian bandits; however, their techniques highly depend on the structure of Gaussian linear bandits and cannot be applied to low-rank \change{matrix or} tensor bandits. This is another interesting future direction. 
}


\section{Simulations}
\label{sec:simulation}

We carry out some preliminary experiments to compare the numerical performance of \texttt{tensor epoch-greedy},  \texttt{tensor elimination} and \texttt{tensor ensemble sampling} with two competitive methods: \texttt{vectorized UCB} which unfolds the tensor into a long vector and then implements standard UCB \citep{auer2002} for multi-armed bandits, and \texttt{matricized ESTR} \citep{jun2019bilinear} which unfolds the tensor into a matrix along an arbitrary mode and implements ESTR for low-rank matrix bandits. 

We first describe the way to generate an order-three true reward tensor ($d=3$) according to Tucker decomposition in \eqref{eq:tucker}. The tensor dimensions are set to be same, i.e., $p_1=p_2=p_3=p$. The triplet of tensor Tucker rank is fixed to be $r_1=r_2=r_3= r =2$. Denote $\tilde{U}_j\in \mathbb{R}^{p_j\times r_j}$ as i.i.d standard Gaussian matrices. Then we apply QR decomposition on $\tilde{U}_j$, and assign the Q part as the singular vectors $U_j$. The core tensor $\cS\in\mathbb{R}^{r\times r\times r}$ is constructed as a diagonal tensor with $\cS_{iii} = wp^{1.5}$, for $1\le i\le r.$ Here, $wp^{1.5}$ indicates the signal strength \citep{zhang2018tensor}. The random noise $\epsilon_t$ is generated i.i.d from a standard Gaussian distribution.

\begin{figure}[h!]
\centering
\begin{tabular}{cc}
(a) $p=15$, $w=0.5$ & (b) $p=20$, $w=0.5$\\
\includegraphics[scale = 0.40]{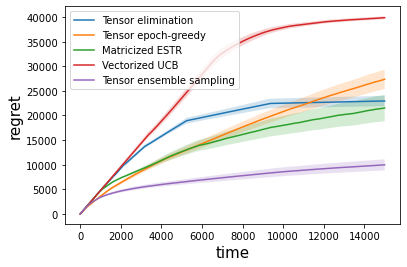} & 
\includegraphics[scale = 0.40]{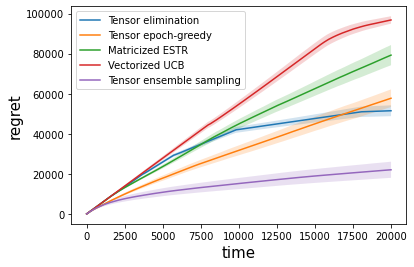}\\
(c) $p = 15$, $w = 0.8$ & (d) $p = 20$, $w=0.8$ \\
\includegraphics[scale = 0.40]{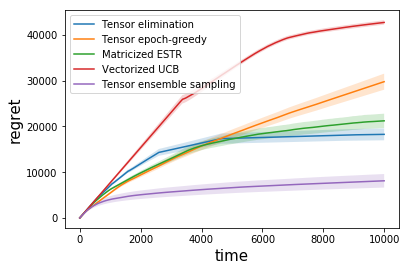}  & 
\includegraphics[scale = 0.40]{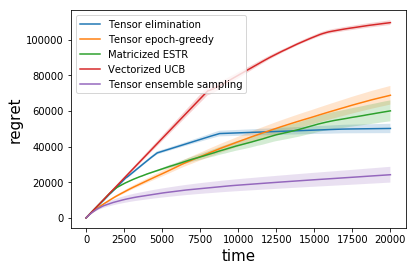}
\end{tabular}
\caption{\change{Cumulative regrets with varying dimension $p$ and signal strength $w$. The shaded areas represent the confidence bands.}}
\label{fig:simulation1}
\end{figure}

\change{
All algorithms involve some hyperparameters, such as the length of initial explorations, width of confidence intervals, the number of rounds of pure explorations and etc. In Section \ref{sec:implementation} of the appendix, we discuss the choice of hyper-parameters for \texttt{tensor elimination}, \texttt{tensor epoch-greedy}, \texttt{tensor ensemble sampling}, and \texttt{matricized ESTR} respectively. In Figure \ref{fig:simulation1}, we report the cumulative regrets of all five algorithms for four settings with $w \in \{0.5, 0.8\}$ and $p \in \{15, 20\}$. All the results are based on 30 replications. Figure \ref{fig:simulation1} shows that \texttt{tensor ensemble sampling} outperforms all other methods in different settings. \texttt{Tensor elimination} does not perform as well as \texttt{tensor ensemble sampling} but is better than other methods for a long time horizon. It aligns with our theoretical findings that \texttt{tensor elimination} has a better overall regret bound for long time horizon, while \texttt{tensor epoch-greedy} is more competitive for small time horizon. When the tensor dimension $p$ increases, the advantage of \texttt{tensor epoch-greedy} in early stage is more apparent. This result 
agrees with our theoretical finding in that 
the regret bound of \texttt{tensor epoch-greedy} has a lower dependency on dimension compared with other methods. 
}

\section{Applications to Online Advertising}
\label{sec:real}

\change{Two real data analysis studies are conducted in the field of online advertising to assess the proposed algorithms. The first study focuses on a contextual tensor bandit problem, while the second study examines a non-contextual tensor bandit problem.}

Our first data set comes from a major internet company and contains the impressions for advertisements displayed on the company's webpages over four weeks in May to June, 2016. The impression is the number of times the advertisement has been displayed. It is a crucial measure to evaluate the effectiveness of an advertisement campaign, and plays an important role in digital advertising pricing. Studying online advertisement recommendation not only brings opportunities for advertisers to increase their ad exposures but also allows them to efficiently study individual-level behavior. 



\begin{figure}[h!]
	\centering
	\includegraphics[scale = 0.2]{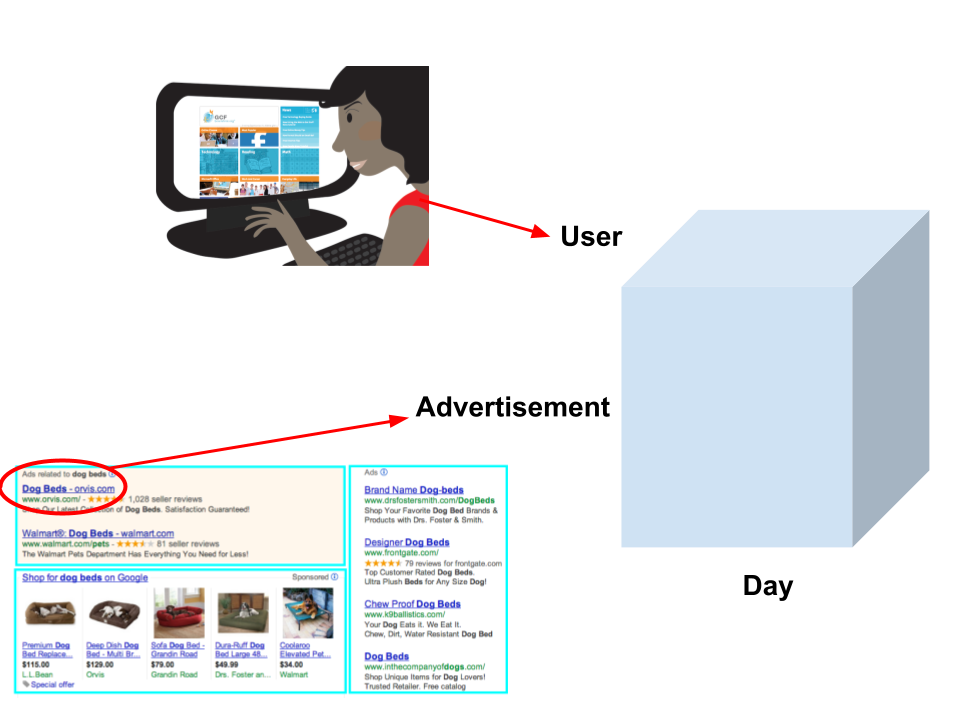}~~
	\includegraphics[scale = 0.5]{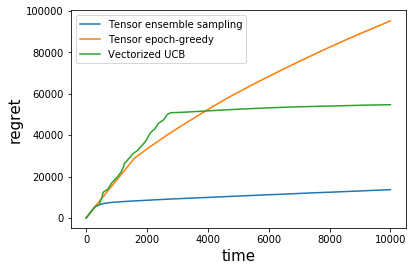}
	\caption{The left plot illustrates the reward tensor formulation in our online advertising data. The right plot shows cumulative regrets of \texttt{tensor epoch-greedy}, \texttt{tensor ensemble sampling} and \texttt{vectorized UCB} in the contextual tensor bandit real data.}
	\label{fig:real}
\end{figure}

The impressions of 20 advertisements were recorded for 20 most active users. In order to understand the user behavior over different days of a week, the data were aggregated by days of a week. Thus, the data forms an order-three tensor of dimension $20\times 7\times 20$ where each entry in the tensor corresponds to the impression for the given combination of user, day of week and advertisement. The goal of this real application is to recommend advertisement to a selected user on a specific day to achieve maximum reward (impression). The user mode and the day-of-week mode are both contextual information and the agent recommends the corresponding optimal advertisement. \texttt{Tensor elimination} and \texttt{matricized ESTR} can only handle the setting where the agent chooses arms without contextual information. \texttt{Tensor epoch-greedy} is for context-free tensor bandits in our theory but it can also be extended to tensor bandit with contextual information. Therefore, we compare the performance of \texttt{tensor epoch-greedy}, \texttt{tensor ensemble sampling} and \texttt{vectorized UCB} in this contextual tensor bandits problem. The cumulative regrets of all these algorithms are shown in Figure \ref{fig:real}.

From the right plot of Figure \ref{fig:real}, we can observe that \texttt{tensor ensemble sampling} achieves the lowest regret for a long time horizon. Comparing \texttt{tensor epoch-greedy} and \texttt{vectorized UCB}, the former is better for a short time horizon. At the last time horizon, \texttt{tensor ensemble sampling} is 75\% lower than that of \texttt{vectorized UCB} and is 85.6\% lower than that of \texttt{tensor epoch-greedy}. The t-test of difference between the mean of final regret for \texttt{tensor epoch-greedy} and \texttt{tensor ensemble sampling} indicates that the two means are significantly different ( t-statistic is 1191.37 and p-value is 0). The t-test between \texttt{tensor ensemble sampling} and \texttt{vectorized UCB} also shows significantly improvement is achieved by \texttt{tensor ensemble sampling} (t-statistic is 1770.33 and p-value is 0). The success of \texttt{tensor ensemble sampling} helps advertisers to better optimize the allocation of ad resources for different users on different days. By tracking users' behavior on ad exposures and conversions over time, advertises can make personalized recommendation based on individual-level data. Besides, our models are maintained and updated based on users' feedback. Such interactive models can be applied to other dynamic and online learning real problems. 

\begin{figure}[h!]
\centering
\includegraphics[scale = 0.70]{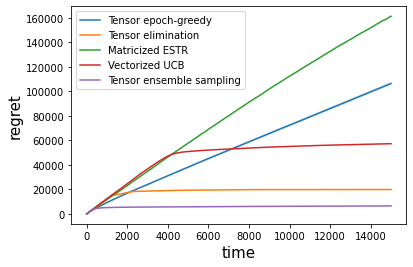}
\caption{\change{Cumulative regrets in non-contextual tensor bandit real data.}}
\label{fig:real_data_add}
\end{figure}

\change{In addition to the aforementioned contextual tensor bandit problem, we further consider a real data analysis on non-contextual tensor bandits. The tensor data used in this study is a third-order tensor that collects information on ad clicks across 20 advertisements, 10 publishers, and 7 days of the week. A {\it publisher} refers to a specific webpage on the online company's website, such as the main homepage, a page dedicated to financial news, or one dedicated to sports news. Each entry in the tensor corresponds to the number of clicks for a particular combination of advertisement, publisher, and day. The goal of this analysis is to identify the optimal combination of advertisement, publisher, and day that results in the highest reward for behavioral targeting purposes \citep{choi2020online, rafieian2021targeting}. For example, if we discover that a particular type of customer prefers ad $i$ on publisher $j$ on day $k$ of the week, we can use this information to target this customer segment in future advertising campaigns by displaying ad $i$ on publisher $j$ on day $k$ of the week to maximize the reward. Since all three modes represent actions, this is a non-contextual tensor bandit problem. 
We conducted a comparison of three proposed algorithms with two baseline models on non-contextual tensor data, and the cumulative regrets of all these algorithms are shown in Figure \ref{fig:real_data_add}. 
It is seen that both \texttt{tensor ensemble sampling} and \texttt{tensor elimination} yielded low regret over a long time horizon, with \texttt{tensor ensemble sampling} performing slightly better. \texttt{Matricized ESTR} has the worst performance. When the time horizon is short, \texttt{tensor epoch greedy} performs better \change{in comparison to \texttt{vectorized UCB}}. These findings are consistent with those from our simulation studies.}

\section*{Acknowledgment}
The authors thank the editor Professor Jane-Ling Wang, the associate editor and two anonymous reviewers for their valuable comments and suggestions which led to a much improved paper. Will Wei Sun's research was partially supported by NSF-SES grant (2217440). Any opinions, findings, and conclusions expressed in this material are those of the authors and do not reflect the views of the National Science Foundation. The authors report there are no competing interests to declare.

\baselineskip=16pt
\bibliographystyle{asa}
\bibliography{ref_tensor}

\begin{thebibliography}{51}
\newcommand{\enquote}[1]{``#1''}
\expandafter\ifx\csname natexlab\endcsname\relax\def\natexlab#1{#1}\fi

\bibitem[{Abbasi-Yadkori et~al.(2011)Abbasi-Yadkori, P{\'a}l, and
  Szepesv{\'a}ri}]{abbasi2011improved}
Abbasi-Yadkori, Y., P{\'a}l, D., and Szepesv{\'a}ri, C. (2011),
  \enquote{Improved algorithms for linear stochastic bandits,} \textit{Advances
  in Neural Information Processing Systems}, 24, 2312--2320.

\bibitem[{Ahn et~al.(2021)Ahn, Kim, and Kang}]{ahn2021accurate}
Ahn, D., Kim, S., and Kang, U. (2021), \enquote{Accurate Online Tensor
  Factorization for Temporal Tensor Streams with Missing Values,} in
  \textit{Proceedings of the 30th ACM International Conference on Information
  \& Knowledge Management}, pp. 2822--2826.

\bibitem[{Allen(2012)}]{Allen12}
Allen, G. (2012), \enquote{Sparse Higher-Order Principal Components Analysis,}
  \textit{Proceedings of the Fifteenth International Conference on Artificial
  Intelligence and Statistics}, 22, 27--36.

\bibitem[{Audibert et~al.(2009)Audibert, Munos, and
  Szepesv{\'a}ri}]{audibert2009exploration}
Audibert, J.-Y., Munos, R., and Szepesv{\'a}ri, C. (2009),
  \enquote{Exploration--exploitation tradeoff using variance estimates in
  multi-armed bandits,} \textit{Theoretical Computer Science}, 410, 1876--1902.

\bibitem[{Auer(2002)}]{auer2002}
Auer, P. (2002), \enquote{Using Confidence Bounds for Exploitation-Exploration
  Trade-offs,} \textit{Journal of Machine Learning Research}, 3, 397--422.

\bibitem[{Auer et~al.(2002)Auer, Cesa-Bianchi, and Fischer}]{auer2002finite}
Auer, P., Cesa-Bianchi, N., and Fischer, P. (2002), \enquote{Finite-time
  analysis of the multiarmed bandit problem,} \textit{Machine learning}, 47,
  235--256.

\bibitem[{Bi et~al.(2022)Bi, Adomavicius, Li, and Qu}]{bi2022improving}
Bi, X., Adomavicius, G., Li, W., and Qu, A. (2022), \enquote{Improving Sales
  Forecasting Accuracy: A Tensor Factorization Approach with Demand Awareness,}
  \textit{INFORMS Journal on Computing}.

\bibitem[{Bi et~al.(2018)Bi, Qu, Shen, et~al.}]{bi2018multilayer}
Bi, X., Qu, A., Shen, X., et~al. (2018), \enquote{Multilayer tensor
  factorization with applications to recommender systems,} \textit{The Annals
  of Statistics}, 46, 3308--3333.

\bibitem[{Bi et~al.(2021)Bi, Tang, Yuan, Zhang, and Qu}]{bi2021tensors}
Bi, X., Tang, X., Yuan, Y., Zhang, Y., and Qu, A. (2021), \enquote{Tensors in
  statistics,} \textit{Annual review of statistics and its application}, 8,
  345--368.

\bibitem[{Bubeck and Cesa-Bianchi(2012)}]{bubeck2012}
Bubeck, S. and Cesa-Bianchi, N. (2012), \enquote{Regret Analysis of Stochastic
  and Nonstochastic Multi-armed Bandit Problems,} \textit{Foundations and
  Trends in Machine Learning}, 5, 1--122.

\bibitem[{Cai et~al.(2021)Cai, Li, Poor, and Chen}]{cai2021nonconvex}
Cai, C., Li, G., Poor, H.~V., and Chen, Y. (2021), \enquote{Nonconvex Low-Rank
  Tensor Completion from Noisy Data,} \textit{Operations Research}.

\bibitem[{Choi et~al.(2020)Choi, Mela, Balseiro, and Leary}]{choi2020online}
Choi, H., Mela, C.~F., Balseiro, S.~R., and Leary, A. (2020), \enquote{Online
  display advertising markets: A literature review and future directions,}
  \textit{Information Systems Research}, 31, 556--575.

\bibitem[{Dwaracherla et~al.(2022)Dwaracherla, Wen, Osband, Lu, Asghari, and
  Van~Roy}]{dwaracherla2022ensembles}
Dwaracherla, V., Wen, Z., Osband, I., Lu, X., Asghari, S.~M., and Van~Roy, B.
  (2022), \enquote{Ensembles for Uncertainty Estimation: Benefits of Prior
  Functions and Bootstrapping,} \textit{arXiv preprint arXiv:2206.03633}.

\bibitem[{Friedland and Lim(2017)}]{friedland2017nuclear}
Friedland, S. and Lim, L.-H. (2017), \enquote{Nuclear norm of higher-order
  tensors,} \textit{Mathematics of Computation}, 87, 1255--1281.

\bibitem[{Frolov and Oseledets(2017)}]{frolov2017tensor}
Frolov, E. and Oseledets, I. (2017), \enquote{Tensor methods and recommender
  systems,} \textit{Wiley Interdisciplinary Reviews: Data Mining and Knowledge
  Discovery}, 7.

\bibitem[{Garivier and Capp{\'e}(2011)}]{garivier2011kl}
Garivier, A. and Capp{\'e}, O. (2011), \enquote{The KL-UCB algorithm for
  bounded stochastic bandits and beyond,} in \textit{Proceedings of the 24th
  annual conference on learning theory}, JMLR Workshop and Conference
  Proceedings, pp. 359--376.

\bibitem[{Han et~al.(2022)Han, Willett, and Zhang}]{han2022optimal}
Han, R., Willett, R., and Zhang, A.~R. (2022), \enquote{An optimal statistical
  and computational framework for generalized tensor estimation,} \textit{The
  Annals of Statistics}, 50, 1--29.

\bibitem[{Id{\'e} et~al.(2022)Id{\'e}, Murugesan, Bouneffouf, and
  Abe}]{ide2022targeted}
Id{\'e}, T., Murugesan, K., Bouneffouf, D., and Abe, N. (2022),
  \enquote{Targeted Advertising on Social Networks Using Online Variational
  Tensor Regression,} \textit{arXiv preprint arXiv:2208.10627}.

\bibitem[{Jain and Oh(2014)}]{jain2014provable}
Jain, P. and Oh, S. (2014), \enquote{Provable tensor factorization with missing
  data,} \textit{Advances in Neural Information Processing Systems},
  1431--1439.

\bibitem[{Jun et~al.(2019)Jun, Willett, Nowak, and Wright}]{jun2019bilinear}
Jun, K., Willett, R., Nowak, R., and Wright, S. (2019), \enquote{Bilinear
  Bandits with Low-Rank Structure,} \textit{36st International Conference on
  Machine Learning}, 97, 3163--3172.

\bibitem[{Katariya et~al.(2017{\natexlab{a}})Katariya, Kveton, Szepesv{\'a}ri,
  Vernade, and Wen}]{katariya2017bernoulli}
Katariya, S., Kveton, B., Szepesv{\'a}ri, C., Vernade, C., and Wen, Z.
  (2017{\natexlab{a}}), \enquote{Bernoulli rank-1 bandits for click feedback,}
  in \textit{Proceedings of the 26th International Joint Conference on
  Artificial Intelligence}, pp. 2001--2007.

\bibitem[{Katariya et~al.(2017{\natexlab{b}})Katariya, Kveton, Szepesvari,
  Vernade, and Wen}]{katariya2017stochastic}
Katariya, S., Kveton, B., Szepesvari, C., Vernade, C., and Wen, Z.
  (2017{\natexlab{b}}), \enquote{Stochastic Rank-1 Bandits,} in
  \textit{Artificial Intelligence and Statistics}, pp. 392--401.

\bibitem[{Kolda and Bader(2009)}]{kolda2009}
Kolda, T. and Bader, B. (2009), \enquote{Tensor Decompositions and
  Applications,} \textit{SIAM Review}, 51, 455--500.

\bibitem[{Kveton et~al.(2017)Kveton, Szepesvari, Rao, Wen, Abbasi-Yadkori, and
  Muthukrishnan}]{kveton2017stochastic}
Kveton, B., Szepesvari, C., Rao, A., Wen, Z., Abbasi-Yadkori, Y., and
  Muthukrishnan, S. (2017), \enquote{Stochastic low-rank bandits,}
  \textit{arXiv preprint arXiv:1712.04644}.

\bibitem[{Kveton et~al.(2020)Kveton, Zaheer, Szepesvari, Li, Ghavamzadeh, and
  Boutilier}]{kveton2020randomized}
Kveton, B., Zaheer, M., Szepesvari, C., Li, L., Ghavamzadeh, M., and Boutilier,
  C. (2020), \enquote{Randomized exploration in generalized linear bandits,} in
  \textit{International Conference on Artificial Intelligence and Statistics},
  PMLR, pp. 2066--2076.

\bibitem[{Lai and Robbins(1985)}]{lai1985asymptotically}
Lai, T.~L. and Robbins, H. (1985), \enquote{Asymptotically efficient adaptive
  allocation rules,} \textit{Advances in applied mathematics}, 6, 4--22.

\bibitem[{Langford and Zhang(2007)}]{langford2007epoch}
Langford, J. and Zhang, T. (2007), \enquote{The epoch-greedy algorithm for
  contextual multi-armed bandits,} in \textit{Proceedings of the 20th
  International Conference on Neural Information Processing Systems}, Citeseer,
  pp. 817--824.

\bibitem[{Lattimore and Szepesv{\'a}ri(2020)}]{lattimore2020bandit}
Lattimore, T. and Szepesv{\'a}ri, C. (2020), \textit{Bandit algorithms},
  Cambridge University Press.

\bibitem[{Li et~al.(2010)Li, Chu, Langford, and Schapire}]{li2010contextual}
Li, L., Chu, W., Langford, J., and Schapire, R.~E. (2010), \enquote{A
  contextual-bandit approach to personalized news article recommendation,} in
  \textit{Proceedings of the 19th international conference on World wide web},
  pp. 661--670.

\bibitem[{Lu and Van~Roy(2017)}]{lu2017}
Lu, X. and Van~Roy, B. (2017), \enquote{Ensemble Sampling,} \textit{Advances in
  Neural Information Processing Systems}.

\bibitem[{Lu et~al.(2018)Lu, Wen, and Kveton}]{lu2018efficient}
Lu, X., Wen, Z., and Kveton, B. (2018), \enquote{Efficient online
  recommendation via low-rank ensemble sampling,} \textit{Proceedings of the
  12th ACM Conference on Recommender Systems}, 460--464.

\bibitem[{Lu et~al.(2021)Lu, Meisami, and Tewari}]{lu2021low}
Lu, Y., Meisami, A., and Tewari, A. (2021), \enquote{Low-rank generalized
  linear bandit problems,} in \textit{International Conference on Artificial
  Intelligence and Statistics}, PMLR, pp. 460--468.

\bibitem[{Negahban and Wainwright(2012)}]{negahban2012restricted}
Negahban, S. and Wainwright, M.~J. (2012), \enquote{Restricted strong convexity
  and weighted matrix completion: Optimal bounds with noise,} \textit{The
  Journal of Machine Learning Research}, 13, 1665--1697.

\bibitem[{Osband et~al.(2018)Osband, Aslanides, and
  Cassirer}]{osband2018randomized}
Osband, I., Aslanides, J., and Cassirer, A. (2018), \enquote{Randomized prior
  functions for deep reinforcement learning,} \textit{Advances in Neural
  Information Processing Systems}, 31.

\bibitem[{Qin et~al.(2022)Qin, Wen, Lu, and Van~Roy}]{NEURIPS2022_874f5e53}
Qin, C., Wen, Z., Lu, X., and Van~Roy, B. (2022), \enquote{An Analysis of
  Ensemble Sampling,} in \textit{Advances in Neural Information Processing
  Systems}, vol.~35.

\bibitem[{Rafieian and Yoganarasimhan(2021)}]{rafieian2021targeting}
Rafieian, O. and Yoganarasimhan, H. (2021), \enquote{Targeting and privacy in
  mobile advertising,} \textit{Marketing Science}, 40, 193--218.

\bibitem[{Richard and Montanari(2014)}]{richard2014statistical}
Richard, E. and Montanari, A. (2014), \enquote{A statistical model for tensor
  PCA,} \textit{Advances in Neural Information Processing Systems}, 2897--2905.

\bibitem[{Russo and Van~Roy(2016)}]{russo2016information}
Russo, D. and Van~Roy, B. (2016), \enquote{An information-theoretic analysis of
  thompson sampling,} \textit{The Journal of Machine Learning Research}, 17,
  2442--2471.

\bibitem[{Russo et~al.(2018)Russo, Van~Roy, Kazerouni, Osband, and
  Wen}]{russo2018tutorial}
Russo, D.~J., Van~Roy, B., Kazerouni, A., Osband, I., and Wen, Z. (2018),
  \enquote{A Tutorial on Thompson Sampling,} \textit{Foundations and
  Trends{\textregistered} in Machine Learning}, 11, 1--96.

\bibitem[{Sen et~al.(2017)Sen, Shanmugam, Kocaoglu, Dimakis, and
  Shakkottai}]{sen2017contextual}
Sen, R., Shanmugam, K., Kocaoglu, M., Dimakis, A., and Shakkottai, S. (2017),
  \enquote{Contextual Bandits with Latent Confounders: An NMF Approach,}
  \textit{Artificial Intelligence and Statistics}, 518--527.

\bibitem[{Song et~al.(2019)Song, Ge, Caverlee, and Hu}]{song2019tensor}
Song, Q., Ge, H., Caverlee, J., and Hu, X. (2019), \enquote{Tensor completion
  algorithms in big data analytics,} \textit{ACM Transactions on Knowledge
  Discovery from Data (TKDD)}, 13, 1--48.

\bibitem[{Sun et~al.(2017)Sun, Lu, Liu, and Cheng}]{Sun2016}
Sun, W.~W., Lu, J., Liu, H., and Cheng, G. (2017), \enquote{Provable sparse
  tensor decomposition,} \textit{Journal of the Royal Statistical Society:
  Series B (Statistical Methodology)}, 79, 899--916.

\bibitem[{Trinh et~al.(2020)Trinh, Kaufmann, Vernade, and
  Combes}]{trinh2019solving}
Trinh, C., Kaufmann, E., Vernade, C., and Combes, R. (2020), \enquote{Solving
  Bernoulli Rank-One Bandits with Unimodal Thompson Sampling,} \textit{31st
  International Conference on Algorithmic Learning Theory}, 1--28.

\bibitem[{Tsybakov(2009)}]{Tsybakov2009}
Tsybakov, A.~B. (2009), \textit{Introduction to Nonparametric Estimation},
  Springer series in statistics, Springer.

\bibitem[{Valko et~al.(2014)Valko, Munos, Kveton, and
  Koc{\'a}k}]{valko2014spectral}
Valko, M., Munos, R., Kveton, B., and Koc{\'a}k, T. (2014), \enquote{Spectral
  bandits for smooth graph functions,} in \textit{International Conference on
  Machine Learning}, pp. 46--54.

\bibitem[{Xia et~al.(2021)Xia, Yuan, and Zhang}]{xia2021statistically}
Xia, D., Yuan, M., and Zhang, C.-H. (2021), \enquote{Statistically optimal and
  computationally efficient low rank tensor completion from noisy entries,}
  \textit{The Annals of Statistics}, 49, 76--99.

\bibitem[{Xu and Yin(2013)}]{xu2013block}
Xu, Y. and Yin, W. (2013), \enquote{A block coordinate descent method for
  regularized multiconvex optimization with applications to nonnegative tensor
  factorization and completion,} \textit{SIAM Journal on imaging sciences}, 6,
  1758--1789.

\bibitem[{Yu et~al.(2015)Yu, Cheng, and Liu}]{yu2015accelerated}
Yu, R., Cheng, D., and Liu, Y. (2015), \enquote{Accelerated online low rank
  tensor learning for multivariate spatiotemporal streams,} in
  \textit{International conference on machine learning}, PMLR, pp. 238--247.

\bibitem[{Yuan and Zhang(2016)}]{yuan2016tensor}
Yuan, M. and Zhang, C.-H. (2016), \enquote{On tensor completion via nuclear
  norm minimization,} \textit{Foundations of Computational Mathematics}, 16,
  1031--1068.

\bibitem[{Zhang and Xia(2018)}]{zhang2018tensor}
Zhang, A. and Xia, D. (2018), \enquote{Tensor SVD: Statistical and
  computational limits,} \textit{IEEE Transactions on Information Theory}, 64,
  7311--7338.

\bibitem[{Zhang et~al.(2019)}]{zhang2019cross}
Zhang, A. et~al. (2019), \enquote{Cross: Efficient low-rank tensor completion,}
  \textit{The Annals of Statistics}, 47, 936--964.

\end{thebibliography}

\newpage
\baselineskip=24pt
\setcounter{page}{1}
\setcounter{equation}{0}
\setcounter{section}{0}
\renewcommand{\thesection}{S.\arabic{section}}
\renewcommand{\thelemma}{S\arabic{lemma}}
\renewcommand{\theequation}{S\arabic{equation}}

\begin{center}
{\Large\bf Supplementary Materials} \\
\medskip
{\Large\bf ``Stochastic Low-rank Tensor Bandits for Multi-dimensional Online Decision Making"}  \\
\bigskip
\vspace{0.5in} 
\end{center}
\bigskip

\noindent

In the appendix, we provide detailed proofs of Theorems  \ref{thm:tensor_eliminator}-\ref{thm:tensor_etc} in Section \ref{sec:proof}, proof of the main lemma in Section \ref{proof:elim}, the equivalent formulation of tensor bandits in Section \ref{sec:equivalence}, and the algorithm for low-rank tensor completion in Section \ref{sec:tensor_alg}. Section \ref{app:ats} contains a general approximate Thompson sampling algorithm and its Bayesian regret bound, and Section \ref{sec:implementation} includes the implementation details of all algorithms in the experiments. 


\section{Proofs of Main Theorems}\label{sec:proof}

\subsection{Proof of Theorem \ref{thm:tensor_eliminator}}\label{proof:tensor_eliminator}
From Lemma \ref{lemma:tensor_completion} and the assumption $\|\cX\|_{\infty}\leq 1$, we know that with probability at least $1-p^{-10}$,
\begin{equation*}
    \big\|\hat{\cX}_{n_1}-\cX\big\|_F\leq C_1\sqrt{\frac{p^{d+1}r\log(p)}{n_1}}.
\end{equation*}
By definitions, $U_i, \hat{\Ub}_i$ are left singular vectors of $\cM_i(\cX)$ and $\cM_i(\hat{\cX}_{n_1})$, respectively. Here, the matricization operator $\cM(\cdot)$ is defined in \eqref{eqn:matri}. Then we can verify
\begin{equation*}
    U_iU_i^{\top}\cM_i(\cX) = U_iU_i^{\top}U_i\Sigma V_i^{\top} = U_i^{\top}\Sigma V_i^{\top} = \cM_i(\cX).
\end{equation*}
Let $\hat{\Ub}_{i\perp}\in\mathbb{R}^{p\times (p - r)}$ be the orthogonal complement of $\hat{\Ub}_i$ for $i\in[d]$. For an orthogonal matrix $U$ and an arbitary matrix $X,Y$, we have $\|UX\|_F\leq \|U\|_2\|X\|_F = \|X\|_F$ and $\|XY\|_F\geq \|X\|\sigma_{\min}(Y)$.   Suppose $\sigma_i$ is the $r$-th singular value of $\cM_i(\cX)$. Using the above fact, we have
\begin{equation*}
    \begin{split}
         &\|\cM_i(\hat{\cX}_{n_1})-\cM_i(\cX)\|_F\\
         \geq& \Vert\hat{\Ub}_{i\perp}^\top(\cM_i(\hat{\cX}_{n_1}) - U_iU_i^\top\cM_i(\cX))\|_F\\
         = &\|\hat{\Ub}_{i\perp}^\top U_iU_i^\top\cM_i(\cX)\|_F\\
    \geq &\|\hat{\Ub}_{i\perp}^\top U_i\|_F\sigma_r(U^\top_i\cM_i(\cX)) = \|\hat{\Ub}_{i\perp}^\top U_i\|_F \sigma_i.
    \end{split}
\end{equation*}
Therefore we have,
\begin{equation}\label{eqn:eigen_bound}
     \|\hat{\Ub}_{i\perp}^\top U_i\|_F\leq \frac{\|\cM_i(\cX)-\cM_i(\hat{\cX}_{n_1})\|_F}{\sigma_i} = \frac{\|\cX-\hat{\cX}_{n_1}\|_F}{\sigma_i}\leq \frac{C_1}{\sigma_i}\sqrt{\frac{p^{d+1}r\log(p)}{n_1}},
\end{equation}
with probability at least $1-p^{-\alpha}$. As discussed in Section \ref{sec:tensor_elimination},  we reformulate original tensor bandits into a stochastic linear bandits with finitely many arms. Recall that $\bbeta=\text{vec}(\cY)$ with
\begin{equation*}
    \cY = \cX \times_1 [\hat{\Ub}_1;\hat{\Ub}_{1\perp}] \cdots \times_d[\hat{\Ub}_d;\hat{\Ub}_{d\perp}]\in\mathbb R^{p_1\times \cdots\times p_d},
\end{equation*}
and the corresponding action set
\begin{eqnarray*}
 \mathbb A := \Big\{\text{vec}\Big([\hat{\Ub}_1;\hat{\Ub}_{1\perp}]^{\top}\be_{\ib_1}\circ \cdots \circ[\hat{\Ub}_d;\hat{\Ub}_{d\perp}]^{\top}\be_{\ib_d} \Big), \ib_1\in[p_1], \ldots, \ib_d\in [p_d]\Big\}.
\end{eqnarray*}
From Eq.~\eqref{eqn:eigen_bound}, we have
\begin{eqnarray}\label{eqn:theta_norm}
\big\|\bbeta_{(q+1):p^d}\big\|_2&\leq&  \prod_{i=1}^d\|\hat{\Ub}_{i\perp}^\top U_i\|_F  \|\cS\|_F\nonumber\\
&\leq& \frac{\|\hat{X}_{n_1}-X\|_F^d}{\Pi_{i=1}^d\sigma_i}\|\cS\|_F\nonumber\\
&\leq&  \frac{r^{d/2}}{\Pi_{i=1}^d\sigma_i} \frac{C_1^dr^{d/2}p^{\frac{d^2+d}{2}}\log^{d/2} (p)}{n_1^{d/2}},
\end{eqnarray}
with probability at least $1-dp^{-\alpha}$. Thus it is equivalent to consider the following linear bandit problem:
\begin{equation*}
    y_t = \langle A_t, \bbeta\rangle+ \epsilon_t,
\end{equation*}
where $\|\bbeta_{(q+1):p^d}\|_2$ satisfies Eq.~\eqref{eqn:theta_norm} and $A_t$ is pulled from action set $\mathbb A$. To better utilize the information coming from low-rank tensor completion, we present the following regret bound for the elimination-based algorithm for stochastic linear bandits with finitely-many arms. The detailed proof is deferred to Section \ref{proof:elim}.
\begin{lemma}\label{thm:regret_low_elim}
Consider the the elimination-based algorithm in Algorithm \ref{alg:low_elim} with $\lambda_{2} = n/(k\log(1+n/\lambda_1))$ and $\lambda_1>0$. With the choice of $\xi= 2\sqrt{14\log(2/\delta)}+\sqrt{\lambda_1} \|\bbeta_{1:q}\|_{2} + \sqrt{\lambda_2}\|\bbeta_{(q+1):p^d}\|_{2}$, the upper bound of cumulative regret of $n$ rounds satisfies 
\begin{equation*}
\begin{split}
      \textrm{R}_n\leq 8\Big(2\sqrt{14\log(2\log (n) p^d/\delta)}+\sqrt{\lambda_1} \|\bbeta_{1:q}\|_{2}\Big)\sqrt{2qn \log(1+\frac{n}{\lambda_1})} + 8\sqrt{2}n\|\bbeta_{(q+1):p^d}\|_{2}.
\end{split}
\end{equation*}
with probability at least $1-\delta$, where $q = p^d-(p-r)^d$.
\end{lemma}

Overall, we can decompose the pseudo regret Eq.~\eqref{eqn:pseduo_regret} into two parts:
\begin{equation*}
    \textrm{R}_n = R_{1n}+ R_{2n} + R_{3n},
\end{equation*}
where $R_{1n}$ quantifies the regret during initialization phase,  $R_{2n}$ quantifies the regret during exploration phase and $R_{3n}$ quantifies the regret during commit phase (linear bandits reduction). Note that $q\leq C_1 p^{d-1}$ for sufficient large $C_1$. Denote 
\begin{equation*}
    \delta_{p,r} =  \frac{r^{d}}{\Pi_{i=1}^{d}\sigma_i}p^{\frac{d^2+d}{2}}\log^{d/2} (p),
\end{equation*}
such that $\|\bbeta_{(q+1):p^d}\|_{2}\leq \delta_{p, r}/n_1^{d/2}$ from Eq.~\eqref{eqn:theta_norm}. Applying the result in Lemma \ref{thm:regret_low_elim} to bound $R_{3n}$ and properly choosing $0<\lambda_1 \leq1/p^d$, we have the following holds with probability at least $1-dp^{-10}-1/n$, 
\begin{equation*}
   \textrm{R}_n\leq C\Big(\underbrace{r^{d/2}p^{d/2}}_{R_{1n}} + \underbrace{n_1}_{R_{2n}}+ \underbrace{\delta_{p,r}n_2/n_1+\sqrt{\log (\log(n_2)) + \log (n_2p^d )}\sqrt{p^{d-1} n_2\log(n_2 p^d)}\Big)}_{R_{3n}},
\end{equation*}
where $n_2 = n-n_1-Cr^{d/2}p^{d/2}$ and $C>0$ is an universal constant. Here, $R_{3n}$ is due to the fact that we run elimination-based algorithm for the rest $n_2$ rounds. For simplicity, we bound all $n_2$ by $n$ as usually did for the proof of explore-then-commit type algorithm.

We optimize with respect to $n_1$ such that 
\begin{equation*}
    n_1 =(n\delta_{p,r})^{\tfrac{2}{d+2}}.
\end{equation*}
It implies the following bound holds with probability at least $1-dp^{-10}-1/n$,
\begin{equation*}
\begin{split}
      \textrm{R}_n&\leq C\Big(r^{d/2}p^{d/2} + \Big(\frac{r^{d}}{\Pi_{i=1}^d\sigma_i}p^{\frac{d^2+d}{2}}\log^{d/2} (p)\Big)^{\tfrac{2}{d+2}}n^{\tfrac{2}{d+2}} \\
      &+\sqrt{\log (\log(n)) + \log (np^d)}\sqrt{p^{d-1} n\log(np^d)}\Big)\\
      &\leq C\Big(r^{d/2}p^{d/2} +\Big(\frac{r^{d}}{\Pi_{i=1}^d\sigma_i}\log^{d/2} (p)\Big)^{\tfrac{2}{d+2}}p^{\tfrac{d^2+d}{d+2}}n^{\tfrac{2}{d+2}} + \sqrt{(d\log(p)+\log(n))^2p^{d-1}n}\Big).
\end{split}
\end{equation*}
This ends the proof.  \hfill $\blacksquare$

\subsection{Proof of Theorem \ref{thm:tensor_etc}}\label{subsec:proof_etc}
The proof uses the trick that couples epoch-greedy algorithm with explore-then-commit algorithm with an optimal tuning. 

\textbf{Step 1.} We decompose the pseudo regret defined in \eqref{eqn:pseduo_regret} as:
\begin{equation*}
    \begin{split}
        \textrm{R}_n &= \sum_{t=1}^n\langle \cA^*-\cA_t, \cX\rangle\\
        & = \sum_{t=1}^{s_1}\langle \cA^*-\cA_{t}, \cX\rangle + \sum_{t=s_1+1}^{n}\langle\cA^*-\cA_{t}, \cX\rangle,
    \end{split}
\end{equation*}
where $s_1$ is the number of initialization steps. After initialization phase, from the definition of exploration time index set in \eqref{def:explor}, the algorithm actually proceeds in phases and each phase contains $(1+ \lceil  s_{2k}\rceil)$ steps: one step random exploration plus $\lceil  s_{2k}\rceil$ steps greedy actions. By algorithm, at phase $k$, the greedy action $\cA_t$ is taken to maximize $\langle\cA_t,\hat{\cX}_{k+s_1}\rangle$ where $\hat{\cX}_{k+s_1}$ is the low-rank tensor completion estimator at phase $k$ based on $(k+s_1)$ random samples. Therefore, we have $\langle \cA_t-\cA^*, \hat{\cX}_{k+s_1}\rangle\geq 0$ and 
\begin{equation*}
    \langle \cA^*-\cA_t, \cX \rangle \leq \langle \cA^*-\cA_t, \cX-\hat{\cX}_{k+s_1}\rangle.
\end{equation*}
By Lemma \ref{lemma:tensor_completion} and the choice of $s_{2k}$ in \eqref{def:number_exploration}, it is sufficient to guarantee 
\begin{equation*}
   \big\|\hat{\cX}_{k+s_1}-\cX\big\|_F \leq 1/s_{2k},
\end{equation*}
holds with probability at least $1-p^{-\alpha}$ from Lemma \ref{lemma:tensor_completion} for any $\alpha>1$. By the Cauchy-Schwarz inequality, we have 
\begin{equation*}
\begin{split}
     \langle \cA^*-\cA_t, \cX \rangle \leq \big\|\cA^*-\cA_t\big\|_F\big\|\hat{\cX}_{k+s_1}-\cX\big\|_F\leq 2/s_{2k},
\end{split}
\end{equation*}
where for the second inequality we use the fact that both tensors $\cA^*$ and $\cA_t$ have only one entry equal to 1 and others are 0. Denote $n_2 = n-s_1$ and $K^* = \min\{K: \sum_{k=1}^K (1+\lceil  s_{2k}\rceil)\geq n_2\}$.  Since we assume $\|\cX\|_{\infty} \leq 1$, the maximum gap $\Delta_{\max}$ is bounded by 2. Then we have
\begin{equation}\label{eqn:regret_bound1}
\begin{split}
     \textrm{R}_n &\leq s_1\Delta_{\max} + \sum_{k=1}^{K^*}\Big(1\cdot \Delta_{\max}+ \lceil  s_{2k}\rceil2/s_{2k}\Big)\\
     &\leq (s_1+K^*)\Delta_{\max} + 2K^* \leq 2s_1 + 4K^*,
\end{split}
\end{equation}
with probability at least $1-K^*p^{-\alpha}$.

\textbf{Step 2.} We  will derive an upper bound for $K^*$. Let $n_2^* = \argmin_{u\in[0, n_2]}[u+(n_2-u)/s_{2u}]$. Consider the following two cases.
\begin{enumerate}
    \item  If $n_2^*\geq K^*$, it is obvious that
$$
K^*\leq n_2^*+(n_2-n_2^*)/s_{2n_2^*}.
$$
\item If $n_2^*\leq K^*-1$, it holds that
\begin{equation*}
    \sum_{k=1}^{K^*-1} s_{2k}\geq \sum_{k=n_2^*}^{K^*-1} s_{2k}\geq (K^*-n_2^*)s_{2n_2^*},
\end{equation*}
where the second inequality is from the fact that $s_{2k}$ is monotone increasing. By the definition of $K^*$, it holds that 
\begin{equation*}
\begin{split}
     n_2-1&\geq \sum_{k=1}^{K^*-1}\big(1+\lceil s_{2k}\rceil \big)\geq \sum_{k=1}^{K^*-1}\big(1+ s_{2k} \big)\geq K^*-1+\big(K^*-n_2^*\big)s_{2n_2^*},
\end{split}
\end{equation*}
which implies 
\begin{equation*}
    K^*\leq n_{2}^*+(n_2-n_2^*)/s_{2n_2^*}.
\end{equation*}
\end{enumerate}
Overall, $K^*$ is upper bounded by $n_{2}^*+(n_2-n_2^*)/s_{2n_2^*}$. 

\textbf{Step 3.} From \eqref{eqn:regret_bound1}, the cumulative regret can be bounded by 
\begin{equation*}
    \textrm{R}_n \leq 2s_1 + 4\min_{u\in[0, n_2]}\Big(u+(n_2-u)/s_{2u}\Big).
\end{equation*}
The second term above is essentially the regret for explore-then-comment type algorithm with the optimal tuning for the length of exploration. Plugging the definition of $s_{2u}$ in \eqref{def:number_exploration} and letting $u=n/s_{2u}$, we have \begin{equation*}
 K^*/2\leq   n_2^* \leq n^{2/3}p^{\tfrac{d+1}{3}}(r\log p)^{\tfrac{1}{3}}.
\end{equation*}	
Thus, we choose $\alpha = \log(2n^{2/3}p^{\tfrac{d+1}{3}}(r\log p)^{\tfrac{1}{3}}p)$ such that $K^*p^{-\alpha}\leq 1/p$. Plugging in $s_1 = C_0r^{d/2}p^{d/2}$, we have \begin{equation*}
   \textrm{R}_n\leq C_0r^{d/2}p^{d/2} + 8\Big(n^{2/3}p^{\tfrac{d+1}{3}}(r\log p)^{\tfrac{1}{3}}\Big) ,
\end{equation*}
with probability at least $1-1/p$. This ends the proof. \hfill $\blacksquare$

\section{Proof of Lemma \ref{thm:regret_low_elim}}\label{proof:elim}


Before we prove it, we introduce some notations first. For a vector $x$ and matrix $V$, we define $\|x\|_V = \sqrt{x^{\top} V x}$ as the weighted $\ell_2$-norm and $\det(V)$ as its determinant. Let $K = \lfloor  \log_2(n)\rfloor$ and $t_k = 2^{k-1}$. Denote $x^* = \argmax_{a\in\mathbb A}\langle a, \bbeta\rangle$.

We have the following regret decomposition by phases:
\begin{equation*}
    \begin{split}
        \textrm{R}_n &= \sum_{t=1}^n \langle x^*-A_t, \bbeta \rangle = \sum_{k=0}^K\sum_{t=t_k}^{t_{k+1}-1}\langle x^*-A_t, \bbeta\rangle\\
        & = \sum_{k=0}^{K}\sum_{t=t_k}^{t_{k+1}-1}\Big(\langle x^*-A_t, \hat{\bbeta}_k\rangle-\langle x^*-A_t, \hat{\bbeta}_k-\bbeta\rangle\Big),
    \end{split}
\end{equation*}
where $\hat{\bbeta}_k$ is the ridge estimator only based on the sample collected in the current phase, defined in Eq.~\eqref{eqn:ridge}. According to Lemma 7 in \citep{valko2014spectral}, for any fixed $x\in\mathbb R^p$ and any $\delta>0$, we have, at phase $k$,
\begin{equation}\label{eqn:confidence_inter}
    \mathbb P\Big(|x^{\top}(\hat{\bbeta}_k-\bbeta)|\leq \|x\|_{V_k^{-1}}\xi\Big)\geq 1-\delta,
\end{equation}
where $\xi= 2\sqrt{14\log(2/\delta)}+\sqrt{\lambda_1} \|\bbeta_{1:q}\|_{2} + \sqrt{\lambda_2}\|\bbeta_{(q+1):p^d}\|_{2}$.
Applying Eq.~\eqref{eqn:confidence_inter} for $x^*$ and $A_t$, we have with probability at least $1-Kp^d\delta$,
\begin{equation*}
    \textrm{R}_n\leq \sum_{k=0}^K\sum_{t=t_k}^{t_{k+1}-1}\langle x^*-A_t, \hat{\bbeta}_k\rangle+\sum_{k=0}^K(t_{k+1}-t_k)\Big(\|x^*\|_{V_k^{-1}}+\|A_t\|_{V_k^{-1}}\Big)\xi.
\end{equation*}
By step \eqref{eqn:elimination} in Algorithm \ref{alg:low_elim}, we have
\begin{equation*}
    \langle x^*-A_t, \hat{\bbeta}_k\rangle\leq \Big(\|x^*\|_{V_k^{-1}}+\|A_t\|_{V_k^{-1}}\Big)\xi.
\end{equation*}

According to Lemma 8 in \cite{valko2014spectral}, for all the actions $x\in \mathbb A_k$ defined in Eq.~\eqref{eqn:elimination},
\begin{equation*}
    \|x\|_{V_k^{-1}}^2 \leq \frac{1}{t_k-t_{k-1}}\sum_{t=t_{k-1}+1}^{t_k}\|x_t\|_{V_{t-1}^{-1}}^2.
\end{equation*}
Then using the elliptical potential lemma (Lemma 19.4 in \cite{lattimore2020bandit}), with probability  at least $1-Kp^d\delta$, we have 
\begin{equation*}
    \begin{split}
        \textrm{R}_n&\leq  2\sum_{k=0}^K(t_{k+1}-t_k) \Big(\|x^*\|_{V_k^{-1}}+\|A_t\|_{V_k^{-1}}\Big)\xi\\
        &\leq4\sum_{k=0}^K (t_{k+1}-t_k)\sqrt{\frac{1}{t_k-t_{k-1}}\log\Big(\frac{\det(V_k)}{\det(\Lambda)}\Big)}\xi,
    \end{split}
\end{equation*}
where $\Lambda = \diag(\lambda_1,\ldots,\lambda_1, \lambda_2, \ldots, \lambda_2)$. According to Lemma 5 in \citep{valko2014spectral}, we have 
\begin{equation*}
   \log \Big(\frac{\det(V_k)}{\det(\Lambda)} \Big)\leq k \log(1+\frac{n}{\lambda_1}) + \sum_{i= k+1}^{p^d}\log (1+\frac{t_i}{\lambda_2}),
\end{equation*}
where $\sum_{i= k+1}^{p^d} t_i\leq T$. With the choice of $\lambda_2$, 
\begin{equation*}
     \log \Big(\frac{\det(V_k)}{\det(\Lambda)} \Big) \leq k \log(1+\frac{n}{\lambda_1}) + \sum_{i= k+1}^{p^d} \frac{t_i}{\lambda_2}\leq 2k \log(1+\frac{n}{\lambda_1}).
\end{equation*}
We know that $t_{k+1}-t_k = 2^{k-1}$ and $t_k-t_{k-1} = 2^{k-2}$. Then one can have 
\begin{equation*}
    \sum_{k=0}^K(t_{k+1}-t_k)\frac{1}{\sqrt{t_k - t_{k-1}}} = \sum_{k=0}^K2^{k/2}\leq \sqrt{n}.
\end{equation*}
Overall, with probability at least $1-Kp^d\delta$, we have
\begin{equation*}
\begin{split}
     \textrm{R}_n&\leq 8\sqrt{2kn \log(1+\frac{n}{\lambda_1})}\Big(2\sqrt{14\log(2/\delta)}+\sqrt{\lambda_1} \|\bbeta_{1:k}\|_{2} + \sqrt{\lambda_2}\|\bbeta_{(k+1):p^d}\|_{2}\Big)\\
     & = 8(2\sqrt{14\log(2\log (n) p^d/\delta)}+\sqrt{\lambda_1} \|\bbeta_{1:k}\|_{2})\sqrt{2kn \log(1+\frac{n}{\lambda_1})} + 8\sqrt{2}n\|\bbeta_{(k+1):p^d}\|_{2}.
\end{split}
\end{equation*}
This ends the proof. \hfill $\blacksquare$

\section{Auxiliary Results}
\label{sec:auxiliary}
\subsection{An equivalent formulation of tensor bandits}\label{sec:equivalence}
We write $\hat{\Ub}_{1\perp}, \ldots,\hat{\Ub}_{d\perp} $ as the orthogonal  basis of the complement subspaces of $\hat{\Ub}_1,\ldots \hat{\Ub}_d$. By definitions, $[\hat{\Ub}_j\hat{\Ub}_{j\perp}]$ is an orthogonal matrix for all $j\in[d]$ such that 
\begin{equation*}
    [\hat{\Ub}_j\hat{\Ub}_{j\perp}][\hat{\Ub}_j\hat{\Ub}_{j\perp}]^{\top} = [\hat{\Ub}_j\hat{\Ub}_{j\perp}]^{\top}[\hat{\Ub}_j\hat{\Ub}_{j\perp}]= \mathbb I_{d\times d}.
\end{equation*}
Denote a rotated true reward tensor as 
\begin{equation*}
    \cY = \cX \times_1 [\hat{\Ub}_1;\hat{\Ub}_{1\perp}] \cdots \times_d[\hat{\Ub}_d;\hat{\Ub}_{d\perp}]\in\mathbb R^{p_1\times \cdots\times p_d},
\end{equation*}
where $\times_1$ is the marginal multiplication defined in Eq.~\eqref{eqn:marginal}. Denote 
$$
\cE_1 = [\hat{\Ub}_1;\hat{\Ub}_{1\perp}]^{\top}\be_{i_{1t}}\circ\cdots \circ [\hat{\Ub}_d;\hat{\Ub}_{d\perp}]^{\top}\be_{i_{dt}}, \cE_2 = \eb_{i_{1t}}\circ\cdots \circ \eb_{i_{dt}}.
$$
We want to prove 
\begin{equation*}
\begin{split}
     &\big\langle \cY,\cE_1\big\rangle =\big\langle \cX, \cE_2\big\rangle.
\end{split}
\end{equation*}
To see this, we use a fact of the Kronecker product (see details in Section 2.6 in \citep{kolda2009}). Let $\cZ_1 \in\mathbb R^{I_1\times \cdots I_N}$ and $A^{(n)}\in\mathbb R^{J_n\times I_n}$ for all $n\in[N]$. Then, for any $n\in[N]$, we have 
\begin{equation*}
\begin{split}
     &\cZ_2 = \cZ_1\times_1A^{(1)}\cdots \times_N A^{(N)} \\
     \Leftrightarrow& \ \cM_{n}(\cZ_2) = A^{(n)}\cM_n(\cZ_1)\Big(A^{(N)}\otimes \ldots \otimes A^{(n+1)}\otimes A^{(n-1)}\otimes \cdots \otimes A^{(1)}\Big)^{\top},
\end{split}
\end{equation*}
where $\cM_{n}(\cZ)$ is the mode-$n$ matricization and $\otimes $ is a Kronecker product. Denote $H = [\hat{\Ub}_2\hat{\Ub}_{2\perp}]\otimes \cdots \otimes [\hat{\Ub}_d;\hat{\Ub}_{d\perp}]$. By a matricization of $\cY, \cE$ along the first mode, we have
\begin{equation*}
    \begin{split}
      \big\langle \cY,\cE_1\big\rangle &= \big\langle  \cM_{1}(\cY), \cM_1(\cE_1) \big\rangle  \\
      &=\big\langle [\hat{\Ub}_1;\hat{\Ub}_{1\perp}] \cM_1(\cX) H^{\top},   [\hat{\Ub}_1;\hat{\Ub}_{1\perp}] \cM_1(\cE_2) H^{\top}  \big\rangle\\
      & =\text{trace}\Big(H\cM_1(\cX)^{\top}[\hat{\Ub}_1;\hat{\Ub}_{1\perp}]^{\top} [\hat{\Ub}_1;\hat{\Ub}_{1\perp}] \cM_1(\cE_2) H^{\top} \Big)\\
      & = \text{trace}\Big(H\cM_1(\cX)^{\top} \cM_1(\cE_2) H^{\top} \Big)\\
      & = \big\langle \cX \times_1 \mathbb I_{d\times d} \times_2 [\hat{\Ub}_2\hat{\Ub}_{2\perp}] \cdots \times_d[\hat{\Ub}_d;\hat{\Ub}_{d\perp}], \be_{\ib_{1t}}\circ\cdots \circ [\hat{\Ub}_d;\hat{\Ub}_{d\perp}]^{\top}\be_{\ib_{dt}}\big\rangle.
    \end{split}
\end{equation*}
Recursively using the above arguments along each mode, we reach our conclusion.

\subsection{Tensor completion algorithm and guarantee} \label{sec:tensor_alg}
For the sake of completeness, we state the tensor completion algorithm in \citep{xia2021statistically}. 
The goal is to estimate the true tensor $\cX\in\mathbb R^{p_1\times \ldots\times p_d}$ from
\begin{equation*}
    y_t = \langle \cX,  \cA_t\rangle + \epsilon_t, t = 1, \ldots, T,
\end{equation*}
where $\cA_t = \be_{\ib_{1t}}\circ\ldots \circ \be_{\ib_{dt}}$. This is a standard tensor completion with uniformly random missing data. The algorithm consists of two stages: spectral initialization and power iteration. 

\textbf{Spectral initialization.} We first construct an unbiased estimator $\cX_{{\rm ini}}$ for $\cX$ as follows:
    \begin{equation*}
        \cX_{\rm ini} = \frac{p_1\cdots p_d}{T}\sum_{t=1}^{n}y_t \cA_t.
    \end{equation*}
For each $j \in[d]$, we construct the following  $U$-statistic:
\begin{equation*}
    \hat{R}_j = \frac{(p_1\cdots p_d)^2}{T(T-1)}\sum_{1\leq t\neq t'\leq T} y_ty_{t'} \cM_j(\cA_t)\cM_j(\cA_t')^{\top},
\end{equation*}
where $\cM_j$ is the mode-$j$ matricization defined in Eq.~\eqref{eqn:matri}. Compute the eigenvectors of $ \{\hat{R}_j\}_{j=1}^d$ with eigenvalues greater than $\delta$, and denote them by $\{\hat{\Ub}_{j}^{(0)}\}_{j=1}^d$.

\textbf{Power iteration.}  Given $\{\hat{\Ub}_{j}^{(l-1)}\}_{j=1}^d$, $\cX_{\rm ini}$ can be denoised via projections to $j$-th mode. For $l = 1, 2, \ldots$, we alternatively update $\{\hat{\Ub}_{j}^{(l-1)}\}_{j=1}^d$ as follows,
	\begin{equation*}
	\begin{split}
	\hat{\Ub}_j^{(l)} =  \text{ first $r_j$ left singular vectors of } \mathcal{M}_j\Big(\cX_{\rm ini}\times_{j'<j} (\hat{\Ub}_{j'}^{(l-1)})^\top \times_{j'>j} (\hat{\Ub}_{j'}^{(l-1)})^\top\Big).
	\end{split}
	\end{equation*}
	The iteration is stopped when either the increment is no more than the tolerance $\varepsilon$, i.e.,
	\begin{equation}
	\begin{split}
	 \left\|\cX_{\rm ini}\times_1 (\hat{\Ub}_1^{(l)})^{\top}\cdots\times_d (\hat{\Ub}_d^{(l)})^{\top}\right\|_{\rm F} - \left\|\cX_{\rm ini}\times_1 (\hat{\Ub}_1^{(l-1)})^{\top} \cdots\times_d (\hat{\Ub}_d^{(l-1)})^{\top}\right\|_{\rm F} \leq \varepsilon, 
	\end{split}
	\end{equation}
	or the maximum number of iterations is reached. With the final estimates $\hat{\Ub}_1, \ldots, \hat{\Ub}_d$, it is natural to estimate $\cS$ and $\cX$ as
	$
	\hat{\cS} = \cX_{\rm ini} \times_1 \hat{\Ub}_1^\top \cdots \times_d \hat{\Ub}_d^\top,\hat{\cX} = \hat{\cS}\times_1 \hat{\Ub}_1 \cdots \times_d \hat{\Ub}_d.
	$
\hfill $\blacksquare$

\begin{lemma}\label{lemma:tensor_completion}
Suppose Assumptions \ref{con:noise}-\ref{con:max} holds. Suppose $\hat{\cX}_T$ is the low-rank tensor estimator constructed from $T$ uniformly random samples by  Algorithm 1 in \cite{xia2021statistically}. Then for any $\alpha>1$, if the number of samples $T\geq C_0\alpha^3 r^{(d-2)/2}p^{d/2}$ for sufficiently large constant $C_0$, the following holds with probability at least $1-p^{-\alpha}$,
	\begin{equation}\label{eqn:tensor_error}
	\frac{\|\hat{\cX}_T-\cX\|_{F}}{\|\cX\|_F} \leq C_1\sqrt{\frac{\alpha rp\log p}{T}},
	\end{equation}
	where $C_1$ is an absolute constant.
\end{lemma}

Lemma \ref{lemma:tensor_completion} is a direct application of Corollary 2 in \cite{xia2021statistically} with some constant terms ignored. 

\change{
\section{Approximate Thompson Sampling and Its Bayesian Regret Bound}
\label{app:ats}

\newcommand{\I}{\mathbb{I}}

This section presents analysis of a broad approximate Thompson sampling (TS) algorithm for tensor bandits. Notably, our \texttt{tensor ensemble sampling} in Algorithm~\ref{alg:ensemble} can be considered as a specific instance of an approximate TS algorithm. The approximate TS algorithm is introduced in Section~\ref{app:ats-alg}, followed by its Bayesian regret bound in Section~\ref{app:ats-regret}. A detailed proof of the regret bound is provided in Section~\ref{app:ats-proof}.


\subsection{Approximate Thompson Sampling}
\label{app:ats-alg}

The general approximate Thompson sampling algorithm is described in Algorithm~\ref{alg:ats}. Note that we define $\cD_{t-1}$ as the ``history" (i.e. the action-reward trajectory) at the beginning of time $t$. Algorithm~\ref{alg:ats} takes two inputs: the first input is a prior distribution $P_0$ over the reward tensor $\cX$, and the second input is an action sampling oracle {\tt sample}.  In particular, the action sampling oracle maps a prior distribution $P_0$ and a ``history" $\cD_{t-1}$ to a probability distribution over the actions. At each time step $t$, Algorithm~\ref{alg:ats} proceeds as follows: it first samples an action $A_t$ based on {\tt sample}, then it pulls arm $A_t$ and receives reward $y_t$, and finally it updates the ``history" based on the action-reward pair $(A_t, y_t)$.

\begin{algorithm}[h!]
\caption{\texttt{Approximate TS for tensor bandits}}
\begin{algorithmic}[1]
\STATE \textbf{Input:} prior $P_0$ over the reward tensor $\cX$, an action sampling oracle ${\tt sample}$
\STATE \textbf{Initialize} $\cD_0 $ as the empty sequence
\FOR{$t=1,2, \cdots$}
\STATE Sample arm $A_t \sim {\tt sample}(\cdot \, | \, P_0, \cD_{t-1})$
\STATE Pull arm $A_t$ and receive reward $y_t$
\STATE Update $\cD_{t} = {\tt append}(\cD_{t-1}, (A_t, y_t))$
\ENDFOR
\end{algorithmic}
\label{alg:ats}
\end{algorithm}

Note that many algorithms can be viewed as a special case of Algorithm~\ref{alg:ats} with a particular choice of ${\tt sample}$. For instance, let $\Pr(A^* \in \cdot | \cD_{t-1})$ denote the posterior distribution over the optimal action $A^*$, then if we choose 
${\tt sample}(\cdot \, | \, P_0, \cD_{t-1}) = \Pr(A^* \in \cdot | \cD_{t-1})$, then Algorithm~\ref{alg:ats} reduces to the standard Thompson sampling algorithm. Hence, Algorithm~\ref{alg:ats} can be viewed as a generalization of the standard Thompson sampling algorithm. Moreover, as we will show later, we can bound the performance of Algorithm~\ref{alg:ats} based on the "distance" between its action sampling distribution ${\tt sample}(\cdot \, | \, P_0, \cD_{t-1})$, and that of the standard Thompson sampling algorithm, $\Pr(A^* \in \cdot | \cD_{t-1})$. That is why it is referred to as the approximate Thompson sampling algorithm. It is also seen that the tensor ensemble sampling algorithm (Algorithm~\ref{alg:ensemble}) can be viewed as another special case of Algorithm~\ref{alg:ats}. In particular, Algorithm~\ref{alg:ensemble} implicitly defines an action sampling function via an ensemble of tensors.

\subsection{Bayes Regret Bound}
\label{app:ats-regret}
We now establish a general regret bound for Algorithm~\ref{alg:ats}. To simplify the exposition, we make the following assumption: 
\begin{assumption}[Bounded reward]
\label{assum:simplifying_analysis}
For all $\cX' \in \mathtt{support}(P_0)$, and all action $\cA$, we assume that
$
y = \langle \cX', \cA \rangle + \epsilon \in [0, 1]
$
with probability $1$.
\end{assumption}
\noindent
Note that this bounded reward assumption is non-essential, and it is assumed to simplify the exposition. In particular, it is satisfied by assuming the noises are sub-Gaussian as in Assumption \ref{con:noise} and the boundedness of tensor $\cX'$ in Assumption \ref{con:max}.

Following the literature in this field \citep{russo2016information, NEURIPS2022_874f5e53}, we develop a Bayes regret bound 
for Algorithm~\ref{alg:ats}. The Bayes regret is defined as
\begin{equation}
\textstyle
\textrm{BR}_n = \mathbb E \left [ R_n \right ] = \mathbb E \left[ \sum_{t=1}^n \left \langle \cX, \cA^* \right \rangle - 
\sum_{t=1}^n \left \langle \cX, \cA_t \right \rangle
\right],
\end{equation}
where the expectation is over the reward tensor $\cX$ under the prior distribution $P_0$. Different from the frequentist regret bound in $(\ref{eqn:pseduo_regret})$, the Bayes regret has an additional expectation over the reward tensor $\cX$.
To simplify the exposition, we define
\begin{equation}
P^*_t (\cdot) = \Pr(A^* \in \cdot | \cD_{t-1}) \quad \text{and} \quad
\bar{P}_t(\cdot) = {\tt sample}(\cdot | P_0, \mathcal{D}_{t-1}).
\end{equation}
Recall that the \emph{Hellinger distance} between $P^*_t$ and $\bar{P}_t$ is defined as
\begin{equation}
\mathbf{d}_{\mathrm{H}}(P^*_t \| \bar{P}_t) = \sqrt{\sum_a \left( \sqrt{P^*_t(a)} - \sqrt{\bar{P}_t(a)}\right)^2}.
\end{equation}
Note that the Hellinger distance is symmetric.
Moreover, Lemma 2.4 in \citet{Tsybakov2009} shows that the Hellinger distance can be bounded by KL divergences in both directions:
for any $P^*_t$ and any $\bar{P}_t$, we have
$
\mathbf{d}^2_{\mathrm{H}}(P^*_t \| \bar{P}_t) \leq \min \left \{ \mathbf{d}_{\mathrm{KL}}(P^*_t \| \bar{P}_t),
\mathbf{d}_{\mathrm{KL}}(\bar{P}_t \| P^*_t)
\right \}.
$
Then we have the following Bayes regret bound for Algorithm~\ref{alg:ats}:
\begin{theorem}
\label{thm:bayes-regret}
Assume that $p_1=p_2=\cdots = p_d = p$, and the bounded reward assumption
(Assumption~\ref{assum:simplifying_analysis}) holds, then under Algorithm~\ref{alg:ats}, we have
\[
\textrm{BR}_n \leq  \sqrt{p^d \, \mathbb{H}(A^*) n / 2} + 2 \sum_{t=1}^n \mathbb{E} \left[\mathbf{d}_{\mathrm{H}} (P_t^* \| \bar{P}_t)\right] ,
\]
where $\mathbb{H}(A^*)$ is the entropy of $A^*$ under the prior distribution.
\end{theorem}
\noindent
The proof for Theorem~\ref{thm:bayes-regret} is provided in Section~\ref{app:ats-proof}.
Note that under the prior distribution $P_0$, $\cX$ is a random variable. Consequently, $A^*$, which is the index of a maximum element\footnote{When there are multiple maximum elements in $\cX$, we assume that there is a fixed tie-breaking rule to choose $A^*$.} of $\cX$, is also a random variable. Since $\mathbb{H}(A^*) \leq d \log p$, Theorem~\ref{thm:bayes-regret} immediately implies that
\[
\textrm{BR}_n \leq \mathcal{\tilde{O}} \left( \sqrt{d p^d n } + \sum_{t=1}^n \mathbb{E} \left[\mathbf{d}_{\mathrm{H}} (P_t^* \| \bar{P}_t)\right] \right).
\]
Finally, note that in the standard Thompson sampling algorithm, we have $P_t^*(\cdot) = \bar{P}_t(\cdot)$, thus, 
$
\textrm{BR}_n \leq \mathcal{\tilde{O}} \left( \sqrt{d p^d n }  \right)
$, which is sublinear in $n$ and matches the regret bound of vectorized UCB. Note that the analysis in Theorem~\ref{thm:bayes-regret} does not exploit the possible low-rank structure of the reward tensor $\cX$. 
Incorporating this low-rank structure into this information-theoretic analysis of approximate Thompson sampling is 
very challenging, and we believe that it will require novel insights on Bayesian inference in low-rank tensors, and possibly additional assumptions on the prior distribution $P_0$. To the best of our knowledge, this issue is not well understood even in the low-rank matrix case. 
This is an interesting but challenging direction for future work.

Theorem~\ref{thm:bayes-regret} is a general Bayes regret bound for approximate Thompson sampling algorithms, based on the quality of the action sampling distribution. To derive an explicit regret bound for the tensor ensemble sampling algorithm (Algorithm~\ref{alg:ensemble}), we need to further bound the Hellinger distance term for the ensemble sampling algorithm. 
This is also challenging and it requires better understanding of how perturbed rewards and priors affect the tensor decomposition (see equation~(\ref{eqn:perturb})), as well as their connections to Bayesian inferences in low-rank tensors. This is another interesting direction for future research. It is worth mentioning that \citet{NEURIPS2022_874f5e53} has provided a Bayes regret bound for ensemble sampling in linear bandits; however, their techniques highly depend on the structure of Gaussian linear bandits and cannot be applied to low-rank matrix or tensor bandits. 


\subsection{Proof for Theorem~\ref{thm:bayes-regret}}
\label{app:ats-proof}

Our proof utilizes the information-theoretic tool in \citet{NEURIPS2022_874f5e53} which provided a Bayes regret bound for ensemble sampling in linear bandits. 


We start by defining some notations. For any time $t$ and any action $a$, define $y_{t,a}$ as the observed reward if the agent takes action $a$ at time $t$. Then, by definition, we have
\[
\textrm{BR}_n = \sum_{t=1}^n \mathbb E \left[ y_{t, A^*} - y_{t, A_t}\right] = 
\sum_{t=1}^n \mathbb E \left[ \mathbb E_t \left[ y_{t, A^*} - y_{t, A_t}\right] \right ],
\]
where $\mathbb E_t [\cdot]$ is a shorthand notation for $\mathbb E [\cdot \, |  \, \mathcal{D}_{t-1}]$. Note that by definition of the Bayes regret, the expectation is also over the reward tensor $\cX$.
Following  Lemma~1 in \citet{NEURIPS2022_874f5e53}, we can decompose $\textrm{BR}_n$ into a ``main regret term" and an ``approximation error term". Specifically, 
for any $t=1,2,\ldots, n$, we have
\[
\mathbb E_t \left[ y_{t, A^*} - y_{t, A_t} \right] = G_t + J_t,
\]
where $G_t$ is the main regret term defined as
\begin{equation}
G_t = \sum_a \sqrt{P_t^*(a) \bar{P}_t(a)} \left( \mathbb E_t \left[ y_{t, a} \middle | A^*=a \right] - \mathbb E_t \left[ y_{t, a} \right] \right)
\end{equation}
and $J_t$ is the ``approximation error  term" defined as
\begin{equation}
J_t = \sum_a \left( \sqrt{P_t^*(a)} - \sqrt{\bar{P}_t(a)} \right) \left( \sqrt{P_t^*(a)} \mathbb E_t \left[ y_{t, a} \middle | A^*=a \right] + \sqrt{\bar{P}_t(a)} \mathbb E_t \left[ y_{t, a} \right] \right)
\end{equation}
The following lemma bounds $G_t$ based on the information gain in $A^*$.
\begin{lemma}
\label{lemma:ats-g_t}
For each time $t=1,2,\ldots, n$, with probability $1$, we have
\[
G_t \leq   \sqrt{p^d  \, \mathbb I_t \left(A^*; (A_t, y_{t, A_t}) \right) /2} ,
\]
where 
$
\mathbb I_t \left(A^*; (A_t, y_{t, A_t}) \right) = \mathbb I \left(A^*; (A_t, y_{t, A_t}) \, \middle | \, \mathcal{D}_{t-1} = \mathcal{D}_{t-1} \right) $
is the conditional mutual information between $A^*$ and $(A_t, y_{t, A_t})$ conditioning on the given history $\mathcal{D}_{t-1}$.
\end{lemma}
\begin{proof}
We follow the proof of Lemma~2 in \citet{NEURIPS2022_874f5e53}. In particular,
\begin{align}
\I_t \left( A^*; (A_t, y_{t, A_t}) \right) \stackrel{(a)}{=} & \, \I_t (A^*; A_t) + \I_t(A^*; y_{t, A_t} | A_t) \nonumber \\
\stackrel{(b)}{=} & \, \I_t(A^*; y_{t, A_t} | A_t) \nonumber \\
= & \, \sum_a \bar{P}_t(a) \I_t(A^*; y_{t, a} | A_t = a) \nonumber \\
\stackrel{(c)}{=} & \, 
\sum_a \bar{P}_t(a) \I_t (A^*; y_{t, a}) \nonumber \\
= & \, \sum_{a, a^*} \bar{P}_t(a) P^*_t(a^*)
\mathbf{d}_{\mathrm{KL}} \left(
\mathrm{Pr}_t (y_{t, a} \in \cdot \, |  \, A^*=a^*) \, \middle \| \,
\mathrm{Pr}_t (y_{t, a} \in \cdot )
\right ),
\end{align}
where (a) follows from the chain rule of mutual information; (b) follows from the fact that $A^*$ and $A_t$ are conditionally independent given $\mathcal{D}_{t-1}$ so that $\I_t (A^*; A_t) = 0$; and (c) follows from the fact that $A_t$ is conditionally independent of $A^*$ and $y_{t, a}$ given $\mathcal{D}_{t-1}$.
Since $y_{t, a} \in [0, 1]$, from Pinsker's inequality, we have
\[
\mathbb{E}_t [y_{t,a} | A^*=a^*] - \mathbb{E}_t [y_{t, a}] \leq \sqrt{\frac{1}{2} \mathbf{d}_{\mathrm{KL}} \left(
\mathrm{Pr}_t (y_{t, a} \in \cdot \, |  \, A^*=a^*) \, \middle \| \,
\mathrm{Pr}_t (y_{t, a} \in \cdot )
\right )}.
\]
Consequently we have
\[
\I_t \left( A^*; (A_t, y_{t, A_t}) \right) \geq 2 \sum_{a, a^*} \bar{P}_t(a) P^*_t(a^*) \left(\mathbb{E}_t [y_{t,a} | A^*=a^*] - \mathbb{E}_t [y_{t, a}] \right)^2.
\]
On the other hand, recall that 
\[
G_t = \sum_a \sqrt{P_t^*(a) \bar{P}_t(a)} \left( \mathbb E_t \left[ y_{t, a} \middle | A^*=a \right] - \mathbb E_t \left[ y_{t, a} \right] \right).
\]
Without loss of generality, we index the actions as $a=1,2,\ldots, p^d$. Following \citet{russo2016information, NEURIPS2022_874f5e53}, we define the $p^d \times p^d$ matrix $\mathbf{M}$ as
\[
\mathbf{M}_{a, a^*} = \sqrt{\bar{P}_t(a) P^*_t(a^*)} \left(\mathbb E_t \left[ y_{t, a} \middle | A^*=a^* \right] - \mathbb E_t \left[ y_{t, a} \right] \right ) 
\]
for all $a, a^* = 1,2,\ldots, p^d$, where $\mathbf{M}_{a, a^*}$ is the $(a, a^*)$-th element in $\mathbf{M}$. Hence,
$G_t = \mathrm{trace}(\mathbf{M})$, while 
$\I_t \left( A^*; (A_t, y_{t, A_t}) \right) \geq 2 \| \mathbf{M} \|_F^2$. Hence, we have
\[
\frac{G_t^2}{\I_t \left( A^*; (A_t, y_{t, A_t}) \right) } \leq \frac{\mathrm{trace}(\mathbf{M})^2}{ 2 \| \mathbf{M} \|_F^2} \stackrel{(a)}{\leq} \frac{\mathrm{rank}(\mathbf{M})}{2} \stackrel{(b)}{\leq} \frac{p^d}{2},
\]
where (a) follows from $\mathrm{trace}(\mathbf{M}) \leq \sqrt{\mathrm{rank}(\mathbf{M})} \| \mathbf{M} \|_F$ (Fact 10 in \citet{russo2016information}), and (b) follows from $\mathrm{rank}(\mathbf{M}) \leq p^d$ since $\mathbf{M}$ is a $p^d \times p^d$ matrix. 
This concludes the proof.
\end{proof}
Based on Lemma~\ref{lemma:ats-g_t}, and following Lemma~3 in \citet{NEURIPS2022_874f5e53}, we can show that 
\[
\sum_{t=1}^n \mathbb E [G_t] \leq \sqrt{p^d \mathbb H(A^*) n / 2} ,
\]
which is based on Cauchy-Schwarz inequality and the chain rule of the mutual information.
Finally, we bound the approximation error term:
\begin{lemma}
\label{lemma:approx-error}
For all $t=1,2,\ldots, n$, we have
\[
\mathbb E \left[ J_t \right] \leq 2 \mathbb{E} \left[\mathbf{d}_{\mathrm{H}} (P_t^* \| \bar{P}_t)\right].
\]
Summing over $t$ gives the second term in the regret bound of Theorem~\ref{thm:bayes-regret}. 
\end{lemma}
\begin{proof}
We follow the proof of Lemma~4 in \citet{NEURIPS2022_874f5e53}.
Recall that
\begin{align}
J_t =& \, \sum_a \left( \sqrt{P_t^*(a)} - \sqrt{\bar{P}_t(a)} \right) \left( \sqrt{P_t^*(a)} \mathbb E_t \left[ y_{t, a} \middle | A^*=a \right] + \sqrt{\bar{P}_t(a)} \mathbb E_t \left[ y_{t, a} \right] \right) \nonumber \\
\stackrel{(a)}{\leq} & \, \sqrt{\sum_a \left( \sqrt{P_t^*(a)} - \sqrt{\bar{P}_t(a)} \right)^2 }  
\left[
\sqrt{ \sum_a P_t^*(a)  \mathbb E_t^2 \left[ y_{t, a} \middle | A^*=a \right]} +
\sqrt{ \sum_a \bar{P}_t(a)  \mathbb E_t^2 \left[ y_{t, a}  \right]}
\right ] \nonumber \\
\stackrel{(b)}{\leq} & \,  \sqrt{\sum_a \left( \sqrt{P_t^*(a)} - \sqrt{\bar{P}_t(a)} \right)^2 } 
\left[
\sqrt{ \sum_a P_t^*(a)} +
\sqrt{ \sum_a \bar{P}_t(a) }
\right ] \nonumber \\
= & \, 2 \mathbf{d}_{\mathrm{H}}(P^*_t \| \bar{P}_t).
\end{align}
Note that inequality (a) follows from the 
Cauchy–Schwarz inequality, and inequality (b) follows from $y_{t, a} \in [0, 1]$. Taking the expectation, we have $\mathbb E \left[ J_t \right] \leq 2 \mathbb{E} \left[\mathbf{d}_{\mathrm{H}} (P_t^* \| \bar{P}_t)\right]$.

\noindent
Since $\sum_{t=1}^n \mathbb E [G_t] \leq \sqrt{p^d \mathbb H(A^*) n / 2}$, and $\mathbb E \left[ J_t \right] \leq 2 \mathbb{E} \left[\mathbf{d}_{\mathrm{H}} (P_t^* \| \bar{P}_t)\right]$, Theorem~\ref{thm:bayes-regret} is proved.
\end{proof}

}

\change{

\section{Implementation Details in Simulations}
\label{sec:implementation}
Before discussing the choices of hyperparameters in the experiments, we would like to mention that parameter tuning in bandit problems is uniquely challenging, as decisions are made in real time and are based on rewards observed from the past. In a bandit environment, once a parameter is used on partial datasets and a decision is made based on it, the regret resulting from that decision is irreversible. Hence, it is not feasible to select hyperparameters using traditional offline methods such as cross validation.

Next we discuss the choice of hyper-parameters for our \texttt{tensor elimination}, \texttt{tensor epoch-greedy}, \texttt{tensor ensemble sampling} in the experiments, and also conducted sensitivity tests on the choice of these parameters. Finally, we discuss how to select hyper-parameters in the competitive methods.

\begin{itemize}
\item The algorithm \texttt{tensor epoch-greedy} has two hyperparameters, a positive constant $C_0$ that determines the length of the initialization phase $s_1=C_0r^{(d-2)/2}p^{(d-2)}$ and a positive constant $C_2$ that determines the length of the exploitation phase $s_{2k}=\left\lceil C_2 p^{-\tfrac{d+1}{2}}r^{-\tfrac{1}{2}}(\log p)^{-\tfrac{1}{2}}(k+s_1)^{\tfrac{1}{2}} \right\rceil$, both are derived in Theorem 2. In our theoretical analysis, the specific choices for $C_0$ and $C_2$ do not affect the order of the derived regret bound. We let $C_0=1$ and found that it gave enough number of steps in the random initialization phase. For $C_2$, we conducted a sensitivity analysis to evaluate the performance of \texttt{tensor epoch-greedy} with regard to varying values of $C_2$. As shown in Figure \ref{fig:parameter_tuning_AE}(a), the regret of \texttt{tensor epoch-greedy} is not sensitive to different values of $C_2$. Hence, we have chosen to fix $C_2=1$ in all numerical experiments. 

\begin{figure}[h!]
\centering
\begin{tabular}{cc}
(a) $\texttt{tensor epoch-greedy}$ parameter $C_2$ & (b) $\texttt{tensor elimination}$ parameter $c_0$\\
\includegraphics[scale = 0.40]{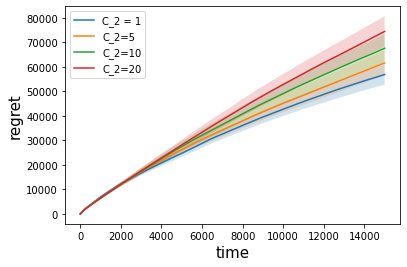} & 
\includegraphics[scale = 0.40]{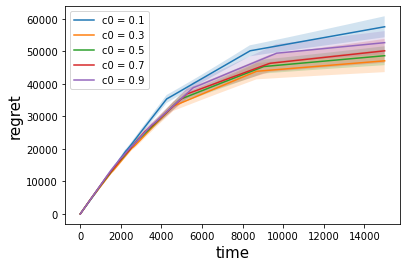}\\
(c) $\texttt{tensor ensemble}$ parameter $\tilde{\sigma}^2$ & \\
\includegraphics[scale = 0.40]{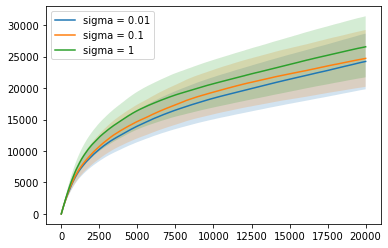}  & 
\end{tabular}
\caption{\change{Top left: Cumulative regrets of different constant multiplier $C_2$ in \texttt{tensor epoch-greedy}. Top right: Cumulative regrets of different exploration length constant multiplier $c_0$ in \texttt{tensor epoch-elimination}. Bottom left: Cumulative regrets of different variance of perturbation noise $\tilde{\sigma}^2$ in \texttt{tensor ensemble sampling}. The shaded areas represent the confidence bands. The simulation setting is same as that in Section 5 with dimension $p_1=p_2=p_3=20$ and $w=0.8$.}}
\label{fig:parameter_tuning_AE}
\end{figure} 

\item The algorithm \texttt{tensor elimination} has one hyperparameter $c_0$ used to determine the number of the exploration steps $c_0n_1$, where $n_1$ follows the theoretical value defined in Theorem 1 and $c_0>0$ is a small constant. For $c_0$, we carried out a sensitivity analysis to evaluate the performance of \texttt{tensor elimination} with regard to varying values of $c_0$. From Figure \ref{fig:parameter_tuning_AE}(b), there is no significant difference between cumulative regrets under different values of $c_0$. Hence, we have chosen to fix $c_0=0.5$ in all numerical experiments.

\item The algorithm \texttt{tensor ensemble sampling} has two hyperparameters including the ensemble size $M$ and the variance of perturbation noise $\tilde{\sigma}^2$. We set $M$ as a relative large number $M=100$ to better approximate posterior distribution and found that it gave a good performance. For $\tilde{\sigma}^2$, we performed a sensitivity analysis to evaluate the performance of \texttt{tensor ensemble sampling} with regard to varying values of $\tilde{\sigma}^2$. As shown in Figure \ref{fig:parameter_tuning_AE}(c), the regret of \texttt{tensor ensemble sampling} is not sensitive to different values of $\tilde{\sigma}^2$. We have chosen to fix $\tilde{\sigma}^2=0.1$ in all numerical experiments.

\end{itemize}

\noindent Similar to \texttt{tensor elimination}, the competitive method \texttt{matricized ESTR} also has a parameter $c_0$ in the initial exploration length. In our experiments, we selected the parameters $c_0\in \{0.1, 0.3, 0.5, 0.7, 0.9\}$ that resulted in the lowest cumulative regret for \texttt{matricized ESTR}, making the comparison favorable to \texttt{matricized ESTR}. Besides, we set the ridge regularization parameter $\lambda_1 = 0.1$ for both \texttt{tensor elimination} and \texttt{matricized ESTR}.

Finally, determining the appropriate rank is still an unresolved issue even in traditional low-rank tensor models, and existing theoretical studies usually assume prior knowledge of the true rank \citep{Sun2016, zhang2018tensor, zhang2019cross, xia2021statistically, cai2021nonconvex, han2022optimal}. In this paper, we adopt this convention and assume prior knowledge of the true ranks for all experiments. However, in practice, one can employ some ad-hoc methods to determine the ranks using uniformly collected samples in the initialization and exploration stages. 

}
\end{document}